\documentclass[pdflatex]{interact}

\usepackage[caption=false]{subfig}
\usepackage[numbers,sort&compress]{natbib}
\usepackage{pdfpages}
\usepackage{amsmath} 
\usepackage{amssymb}  
\usepackage{amsfonts}
\usepackage{bm}
\usepackage{color}
\usepackage{enumerate}
\usepackage{diagbox}
\usepackage{newtxtext,newtxmath}
\usepackage{algpseudocode}
\usepackage{algorithm}
\usepackage{comment}
\usepackage{here}

\bibpunct[, ]{[}{]}{,}{n}{,}{,}
\renewcommand\bibfont{\fontsize{10}{12}\selectfont}
\makeatletter
\def\NAT@def@citea{\def\@citea{\NAT@separator}}
\makeatother

\theoremstyle{plain}

\theoremstyle{definition}

\theoremstyle{remark}

\newcommand{\argmax}{\mathop{\rm arg~max}\limits}
\newcommand{\argmin}{\mathop{\rm arg~min}\limits}

\DeclareMathOperator{\myend}{end}
\usepackage{url}

\begin{document}

\articletype{ARTICLE TEMPLATE}

\title{Route Design in Sheepdog System--Traveling Salesman Problem Formulation and Evolutionary Computation Solution--}

\author{
\name{W. Imahayashi\textsuperscript{a}\thanks{CONTACT W. Imahayashi. Email: imahayashi.wataru@nihon-u.ac.jp}, Y. Tsunoda\textsuperscript{b}, and M. Ogura\textsuperscript{c}}
\affil{\textsuperscript{a}College of Engineering, Department of Mechanical Engineering, Nihon University, Fukushima, Japan; \textsuperscript{b}Graduate School of Engineering, Osaka University, Osaka, Japan; \textsuperscript{c}Graduate School of Information Science and Technology, Osaka University, Osaka, Japan}
}

\maketitle

\begin{abstract}
In this study, we consider the guidance control problem of the sheepdog system, which involves the guidance of the flock using the characteristics of the sheepdog and sheep.
Sheepdog systems require a strategy to guide sheep agents to a target value using a small number of sheepdog agents, and various methods have been proposed.
Previous studies have proposed a guidance control law to guide a herd of sheep reliably, but the movement distance of a sheepdog required for guidance has not been considered.
Therefore, in this study, we propose a novel guidance algorithm in which a supposedly efficient route for guiding a flock of sheep is designed via  Traveling Salesman Problem and evolutionary computation.
Numerical simulations were performed to confirm whether sheep flocks could be guided and controlled using the obtained guidance routes.
We specifically revealed that the proposed method reduces both the guidance failure rate and the guidance distance.
\end{abstract}


\section{Introduction}
Recent years have seen the emergence of multi-agent systems, where multiple agents work cooperatively and interact with each other to accomplish a given task~\cite{Notomista,Sugano,Ikeda,Cai,Atman,Kyo}.
This system is scalable and flexible, and the entire system's performance can potentially improve by increasing the number of agents.
In addition, this system has fault-tolerant performance, which helps to maintain the efficiency of the system even in the event of agent failure.
The development of this system with these advantages is significant and is expected to be applied in various fields~\cite{Notomista,Ikeda,Sugano,Cai,Atman,Tsunoda}.
However, as the number of agents increases, the number of controllers required also increases, making the design of the overall system more challenging.

This study focuses on sheepdog systems~\cite{shepherding_survey,information sensor,Shepherding_PSO,Shepherding_RL,Flock_control,strombom,Tsunoda,Tsunoda_near,Tsunoda_near_kai,Tsunoda_ex,Himo,Fujioka,Tsunoda_theoretical,Tsunoda_ex2}, which are multi-agent systems that can be operated by simply designing controllers for a few agents.
In a sheepdog system, sheep agents perform a given task using a small number of sheepdog agents.
Further, a small number of sheepdogs can control the behavior of the flock against the collective behavior of the majority of sheep.
In this context, the main objective of research on sheepdog systems is to construct a guidance and control law for sheepdogs to guide a flock to its destination.

Related studies on sheepdog system guidance and control laws have been reported, including experiments on theoretical analyses~\cite{Tsunoda,Tsunoda_near,Tsunoda_near_kai,strombom,Fujioka,Himo,Tsunoda_theoretical} and actual equipment~\cite{Flock_control,Tsunoda_ex,Tsunoda_ex2,pierson}.
For example, Vaughan et al.~\cite{Flock_control} achieved the guidance and control of a mobile robot to a destination by mathematically modeling the behavior of a flock of sheep and sheepdogs, with the mobile robot as the sheepdog and the ducks as the flock of sheep. Tsunoda et al.~\cite{Tsunoda_ex} constructed a sheepdog system using a small mobile robot and proposed a framework for actual guidance and control. Also, Tsunoda et al.~\cite{Tsunoda,Tsunoda_near,Tsunoda_near_kai} proposed a control law to guide sheepdogs to track the sheep that are farthest from their destination and showed a high success rate in numerical simulations. Furthermore, they have proposed a guidance control law that focuses on the nearest sheep from the sheepdog and the sheep's velocity vector.

\begin{figure*}[t]
 \centering
  \includegraphics[clip,scale=0.42]{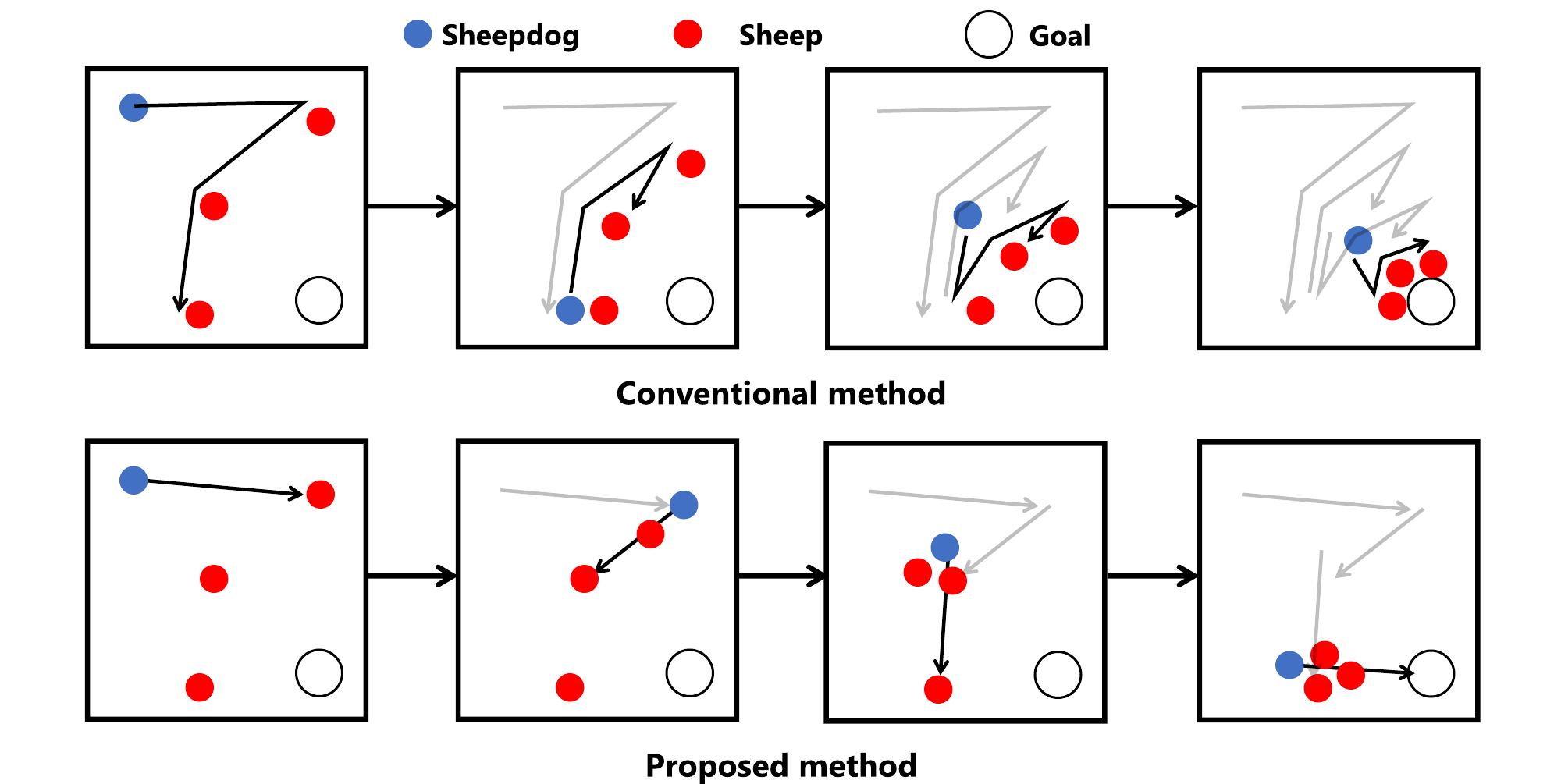}
  \caption{Guidance strategy (upper: conventional, lower: proposed)}
  \label{guidance_explanation}
\end{figure*}

Strömbom et al.~\cite{strombom} analyzed the time series data of an actual sheep flock and sheepdog positions and suggested that sheepdogs use two movements: one to gather sheep and the other to guide sheep. They proposed a guidance control law for these two movements and demonstrated its effectiveness through simulations.
Fujioka et al.~\cite{Fujioka} and Himo et al.~\cite{Himo} proposed guidance control law for sheep with different dynamics.
The former utilized the nominal dynamics model of sheep to distinguish sheep with different behaviors and proposed control rules to guide only the sheep with the nominal dynamics model. The latter proposed control rules to lead all sheep to the destination when some sheep do not show escape behavior from the sheepdog.

The main focus of the aforementioned studies is to develop control laws that reliably guide sheep to their destination. Few studies have considered the distance traveled by sheepdogs to guide them. Considering that the total distance that a sheepdog can travel in one guidance mode is limited and considering guidance control using a robot, as in related studies~\cite{Flock_control,Tsunoda_ex}, it is essential to design a control law that considers the travel distance from the viewpoint of the capacity of the energy source of the robot and the fault tolerance of the hardware.

Therefore, this study aims to propose a strategy for guiding and controlling a sheepdog system to its destination, ensuring the certainty of guidance (guidance success rate) and reducing the total distance traveled by the sheepdog.
This is in contrast to the strategy proposed by Tsunoda et al.~\cite{Tsunoda,Tsunoda_near,Tsunoda_near_kai} (upper part of Figure~\ref{guidance_explanation}) in which sheep are followed as far as possible from the destination. 
Specifically, as shown in the lower part of Figure~\ref{guidance_explanation}, the shepherd repeatedly leads the sheep to the destination and collects them once they have been driven there. After all the sheep have gathered under the sheepdog, the sheepdog leads the flock to its actual destination.
In this guidance strategy, the total moving distance required for guidance changes depending on which sheep the sheepdog visits. Therefore, it is treated as a combinatorial optimization problem (Traveling Salesman Problem) with the moving distance of a sheepdog as the cost, and seeking a guidance route that reduces this cost.
In contrast to previous studies in this field, our research presents a novel approach by determining the order in which the sheep are guided. 
This distinctive method addresses the guidance distance required for a sheepdog to sheep, offering a fresh perspective on the sheepdog system. 
This unique perspective sheds light on whether it is possible to reduce the guiding distance required for a sheepdog to guide a sheep, ultimately paving the way for energy efficiency.
Our research introduces a pioneering framework that not only advances the current understanding of the sheepdog system but also provides a valuable foundation for future investigations in the applicability of the sheepdog system in a swarm robot system.

The remainder of this study is organized as follows. In Section~2, the sheepdog system and control objectives are described. In Section~3, the proposed method is explained. In Section~4, numerical simulations are performed to confirm the effectiveness of the proposed method. In Section~5, the results are summarized.

\section{Problem statement}

This section describes the modeling and control objectives of the sheepdog system.
In this study, we consider an agent, called sheepdog, and $N$ agents, called sheep, moving in a two-dimensional plane~$ \mathbb{R}^{2}$.
We first describe the behavioral rules of sheep. We then state the control objectives and the behavioral rules of sheepdogs. 

\subsection{Sheep Behavior Rules}

We first describe the dynamics of the sheep agents. 
Our sheep behavior rules were based on Reynolds et al.~\cite{Boidmodel} to model the herding behavior of autonomous sheep (i.e., collision avoidance, alignment, and cohesion), to which we add the dynamics of fleeing  sheepdogs.
Let $\mathbb{R}$ denote the set of real numbers, $\mathbb{Z}$ the set of integers, $\mathbb{N}$ the set of natural numbers, and $\mathbb{R_{+}}$ the set of positive real numbers, and the Euclidean norm of $x \in \mathbb{R}^{2}$ is denoted as $\| x \|$. For the set~$\mathcal{N}$, the number of elements is denoted as $\lvert \mathcal{N} \rvert$.
Let $x_{i}[k] \in \mathbb{R}^2$ denote the position of the sheep~$i$ ($i=1, \dotsc, N$) at time~$k$.
We assume that sheep~$i$ is subjected to each of the above three forces by another sheep in a closed disk of radius $r_{s} \in \mathbb{R}_{+}$, i.e., those having the index in the set 
\begin{align}
\mathcal{S}_{i}[k]=\{ \eta \in \{ 1, \ldots, N \} \backslash \{i\} \mid \| x_{i}[k] - x_{\eta}[k] \| \leq r_{s}\} .
\end{align}
Moreover, sheep~$i$ is subjected to a repulsive force depending on its distance from the sheepdog.
Specifically, the time evolution of position~$x_{i}[k]$ of sheep~$i$ at time~$k \in \mathbb{N}$ is given by
\begin{align}
x_{i}[k + 1] = x_{i}[k] + v_{i}[k] \label{eq:position}
\end{align}
for velocity vector
\begin{align}
v_{i}[k] = K_{s_{1}}v_{i, 1}[k] + K_{s_{2}}v_{i, 2}[k] + K_{s_{3}}v_{i, 3}[k] + K_{s_{4}}v_{i, 4}[k], \label{eq:sheep_velo}
\end{align}
where $K_{s_{1}}$, $K_{s_{2}}$, $K_{s_{3}}$, $K_{s_{4}} \geq 0$ are the gain constants and assumed to be the same for all sheep.
Furthermore, $v_{i, l}$ ($l=1,2,3$) corresponds to collision avoidance, alignment, and cohesion in the Boid model, respectively, and are given by 
\begin{align}
v_{i, 1}[k] &= -\frac{1}{|\mathcal{S}_{i}[k]|} \sum_{\eta \in \mathcal{S}_{i}[k]} \frac{x_{\eta}[k] - x_{i}[k]}{\| x_{\eta}[k] - x_{i}[k] \|^{3}},\\
v_{i, 2}[k] &=   \frac{1}{|\mathcal{S}_{i}[k]|} \sum_{\eta \in \mathcal{S}_{i}[k]} \frac{v_{\eta}[k-1]}{\| v_{\eta}[k-1]\|},\\
v_{i, 3}[k] &=   \frac{1}{|\mathcal{S}_{i}[k]|} \sum_{\eta \in \mathcal{S}_{i}[k]} \frac{x_{\eta}[k] - x_{i}[k]}{\| x_{\eta}[k] - x_{i}[k] \|}, 
\end{align}
while 
$v_{i, 4}$ is the repulsion from the sheepdog and is expressed as 
\begin{align}
v_{i, 4}[k] = -\frac{x_{d}[k] - x_{i}[k]}{\| x_{d}[k] - x_{i}[k] \|^{3}}, 
\end{align}
in which $x_{d}[k] \in \mathbb{R}^2$ represents the position of the sheepdog at time~$k$. 
Hence, sheep flee from the shepherd because of the repulsion of the shepherd due to the final term $v_{i, 4}$. In our sheepdog system, the sheepdog uses the escape behavior and the tendency to act as a flock to control sheep for successful guidance of sheep.

\subsection{Control objective}

In this paper, our minimum control requirement to of the sheepdog system is to drive all the sheep positions into a closed disk having radius~$g_{r}$ and centered at~$x_{g} \in \mathbb{R}^2$ (called destination position) in finite time, that is, to realize
\begin{align}\label{eq:goaled}
\|x_{i}[k] - x_{g}\| \leq g_{r} 
\end{align}
for all $i=1, \dotsc, N$. 
Building upon the requirement, our control objective in this research is to reduce the total travel distance taken by the sheepdog to gather the flock defined by 
\begin{align}
J[k_{{\myend}}] = \sum_{k=1}^{k_{{\myend}}} \| x_{d}[k]- x_{d}[k-1] \|, \label{dog_JK}
\end{align}
where $k_{\myend}$ denotes the minimum $k$ such that \eqref{eq:goaled} is satisfied for all $i =1, \dotsc, N$.

\subsection{Farthest Agent Targeting method} \label{tsunoda_method}

In this study, we used the Farthest Agent Targeting (FAT) method proposed by Tsunoda et al.~\cite{Tsunoda_near,Tsunoda_near_kai}. Below we briefly describe the algorithm. 
First, from time~$k$ to $k+1$, the position of the sheepdog is updated as 
\begin{align}
x_{d}[k + 1] = x_{d}[k] + v_{d}[k], \label{dog_x}
\end{align}
where $v_{d}[k]$ is the velocity vector of the sheepdog.
Let $t[k]$ be the index of the sheep farthest from the destination at times $k$, and $n[k]$ be the index of the sheep closest to the sheepdog at time~$k$.
Then, the control law for the sheepdog is given by 
\begin{align}
v_{d}[k] = K_{d_{1}}v_{d_{1}}[k] + K_{d_{2}}v_{d_{2}}[k] + K_{d_{3}}v_{d_{3}}[k], \label{v_d}
\end{align}
where $K_{d_{1}}, K_{d_{2}}, K_{d_{3}} \geq 0$ are constants.
In addition, $v_{d_{i}}[k]~(i=1,2,3)$ are described by the following equations:
\begin{align}
v_{d_{1}}[k] &= -\frac{x_{d}[k] - x_{t{[k]}}[k]}{\| x_{d}[k] - x_{t{[k]}}[k] \|},      \label{eq:dog1}\\
v_{d_{2}}[k] &= \frac{x_{d}[k] - x_{n{[k]}}[k]}{\| x_{d}[k] - x_{n{[k]}}[k] \|^{3}},  \label{eq:dog2}\\
v_{d_{3}}[k] &= \frac{x_{d}[k] - x_{g}}{\| x_{d}[k] - x_{g} \|},  \label{dog3}
\end{align}
representing the gravitational forces acting on the sheepdog
on the farthest sheep
\begin{align}
t[k] &= \argmax_{i \in \{ 1, \ldots, N \} } \| x_{g} - x_{i}[k]\|
\end{align}
from the target, the repulsive force on the closest sheep
\begin{align}
n[k] &= \argmin_{i \in \{ 1, \ldots, N \} } \| x_{d}[k] - x_{i}[k]\|
\end{align}
to the sheepdog, and the repulsive force between the sheepdog and destination position~$x_{g}$, respectively.

\section{Strategies for guiding sheepdogs based on the traveling salesman problem}

\begin{figure*}[t]
 \centering
  \includegraphics[clip,scale=0.42]{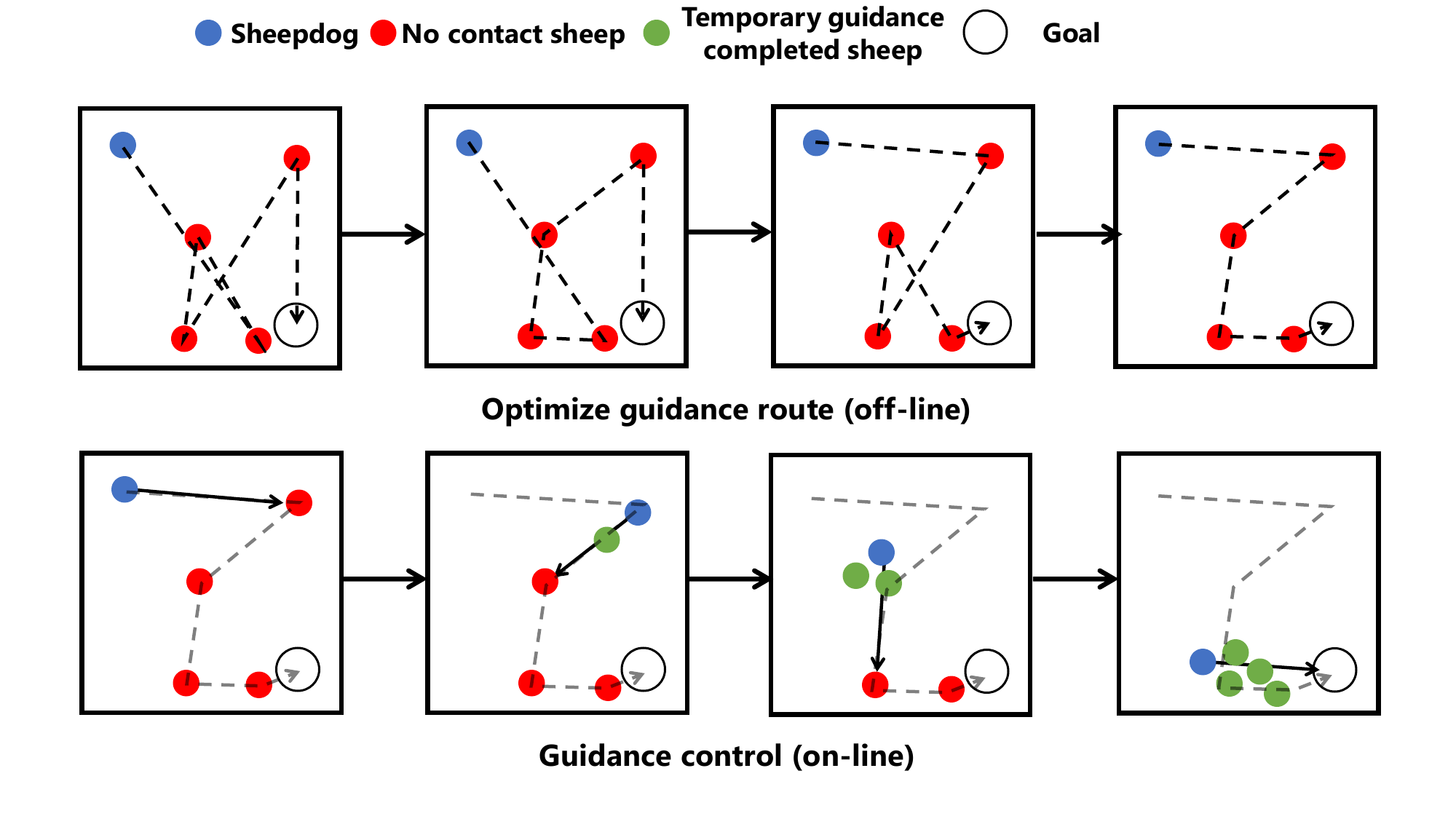}
  \vspace{-8mm}
  \caption{Proposed guidance control strategy}
  \label{proposed_guidance_explanation}
\end{figure*}

In this section, we describe the proposed algorithm for shepherding. In the proposed method, the first-stage part (upper part of Figure~\ref{proposed_guidance_explanation}) should be designed before guidance, and the second-stage part (lower part of Figure~\ref{proposed_guidance_explanation}) should be designed during guidance.
The following methods are used for guidance:
\begin{enumerate}[1]
   \item For the initial locations of both the sheepdog and flock, we computationally determine the distances the sheep will traverse during a single passage. Subsequently, we devise a ``sheep tour'' that effectively minimizes the overall traversal distance (upper part of Figure~\ref{proposed_guidance_explanation}).
   \item The sheepdog moves according to the route of the sheep determined in the first step and drives the sheep to the destination (lower part of Figure~\ref{proposed_guidance_explanation}).
\end{enumerate}

We first in Section~\ref{subsec3.1} describe a combinatorial optimization problem for designing the sheep tour using the Traveling Salesman Problem (TSP) and Randomized Local Search (RLS). 
In Section~\ref{subsec3.2}, we then explain the guided control law for the sheepdog to drive the sheep based on the determined sheep-rounding method using previous studies as support.

\subsection{Determining the order of guidance by TSP}\label{subsec3.1}

If we consider the initial position of the sheepdog as the starting point and the destination as the endpoint, 
the cost of $W(\pi)$ (called the guidance cost) for each round of sheep is
\begin{align} \label{zyunkai}
W(\pi) = \sum_{i=1}^{N} r_{\pi_{i}},
\end{align}
where $\pi$ is a $N$-element finite sequence of sheep indices. We let $\pi_{i}$ denote the $i$-th element of~$\pi$.
Let us define

\begin{align}
    r_{\pi_{i}}=
    \begin{cases}
        \| x_{d}[0] -x_{\pi_{1}}[0]             \|, & \mbox{if $i=1$,}\\
        \| x_{\pi_{i}}[0] - x_{\pi_{i+1}}[0]    \|, & \mbox{if $i\neq 1$ and $i\neq N$,}\\
        \| x_{\pi_{N}}[0] - x_{g}                \|. & \mbox{otherwise.}
    \end{cases}
\end{align}

By reducing $W(\pi)$ given in Equation \eqref{zyunkai}, the total distance traveled by the sheepdog can be expected to be reduced according to the guidance control law described in Section~\ref{subsec3.2}.
To reduce $W({\pi})$, the elements of the finite sequence~$\pi$ should be swapped, which is equivalent to manipulating the order in which the sheepdog moves around the sheep. This type of problem can be attributed to a famous combinatorial optimization problem called the Traveling Salesman Problem.

The Traveling Salesman Problem belongs to a class of problems called NP-hard~\cite{TSP},
where the number of elements in the set significantly increases the computational complexity.
Therefore, to avoid this problem, this study aims to reduce the guidance cost using a Randomized local search (RLS)~\cite{RLS}, an evolutionary computation method, to replace the element order of a finite sequence~$\pi$.

The operating algorithm for the RLS method is as follows:
\begin{enumerate}[1]
   \item Generate a finite sequence~$\pi$ at random.
   \item The operation of reordering elements generates a finite sequence~$\pi^{\prime}$ from a finite sequence~$\pi$.
   \item If $W(\pi^{\prime}) \leq W(\pi)$ holds, then set~$\pi = \pi^{\prime}$.
   \item Repeat 2--3 until the termination condition is met.
\end{enumerate}
Major reordering strategies~\cite{RLS_metamol} in the 2nd step are Exchange, Jump, and Reverse, and we use these three strategies later in our numerical simulation for evaluating the proposed method. 
The RLS method is expected to yield a finite sequence~$\pi$ such that the guidance cost is low. According to the obtained finite sequence, the sheepdog drives the sheep to the destination using the guidance control law described in the next section.

\subsection{Guidance and control law for sheepdog based on finite sequences}\label{subsec3.2}

In this guidance control law, all the sheep are guided to their destinations using a finite sequence~$\pi$ with small guidance cost $W(\pi)$ obtained in Section~\ref{subsec3.1}, whereas the sheepdog moves around the sheep in turn, increasing the number of sheep to be guided. To implement this control law, we use the method proposed by Himo et al.

Specifically, the proposed guidance control law consists of the following actions:
\begin{enumerate}[A]
\item 
For each initial position of the sheepdog and the sheep flock as well as the destination, we generate a path passing through each of the $N+1$ locations only once and minimizing its total path length (see the upper half of Fig.~2 for an illustration).
\item The sheepdog guides all the sheep to the destinations by following the path generated in the previous step. 
\end{enumerate}
Here, temporary destination refers to the location of the sheep, which is defined separately from the destination~$s_g$.
Provisional directing refers to directing the sheep toward a provisional destination.
In Operation~A, the sheep that have already been provisionally directed are collected at the provisional destination. This operation is repeated for all sheep.
After completing Operation~A, the system moves to Operation~B to guide all sheep to their destination (see bottom half of Figure~\ref{proposed_guidance_explanation}).
The details of the guidance rules are explained below.

\begin{algorithm}[tb]
\caption{Proposed Algorithm}
    \label{flowchart}
    \begin{algorithmic}[1]
        \Procedure{Proposed Algorithm}{}
            \State{\textbf{Initialize:}}
            \State{$\mathcal{I}[0]\leftarrow \{\varnothing \}$}
            \State{$\pi_{{\rm ini}} \leftarrow$ generate from uniform random numbers}
            \State{$\pi_{{\rm opt}} \leftarrow$RLS$(\pi_{{\rm ini}}$,\textbf{algorithm},iter)}
            \State{$\nu \leftarrow 1$}
            
            \For{$k \leftarrow 1,2,\ldots,T$}
                \If{$\nu = 1$}
            				\If{$\| x_{\pi_{{\rm opt},1}}[k]-x_{d}[k]\| \leq r_{d}$}
            						\State{append $\pi_{{\rm opt},1}$ to $\mathcal{I}[k]$}
            						\State{$\nu \leftarrow \nu + 1$}
							\EndIf
					\Else
							\If{$\| x_{\pi_{{\rm opt},\nu}}[k]-x_{i}[k]\| \leq g_{r}$ for all $i \in \mathcal{I}[k]$}
            						\State{append $\pi_{{\rm opt},\nu}$ to $\mathcal{I}[k]$}
            						\State{$\nu \leftarrow \nu + 1$}
							\EndIf
					\EndIf
            		\If{$\nu < N$}
            				\State{Sheepdog Update($\mathcal{I}[k]$,$x_{\pi_{{\rm opt}},\nu}[k]$)}
            		\Else
            				\State{Sheepdog Update($\mathcal{U}$,$x_{g}$)}
            		\EndIf
            \EndFor
            
        \EndProcedure
        \Procedure{TSP}{$\pi,w(\pi_{i}),N$}
            \State{Return $W(\pi)\leftarrow \sum_{i=1}^{N} r_{\pi_{i}}$}
        \EndProcedure
        \Procedure{RLS}{$\pi_{0}$,\textbf{algorithm},iter}
        		\State{$\pi \leftarrow \pi_{0}$}
             \For{$j \leftarrow 1,2, \ldots \, $iter}
        				\State{$\pi^{\prime} \leftarrow$update finite sequence~$\pi$ by \textbf{ algorithm}}
        				\If{TSP$(\pi^{\prime},w(\pi_{i}^{\prime}),N) \leq$ TSP$(\pi,w(\pi_{i}),N)$}
        						\State{$\pi \leftarrow \pi^{\prime}$}
        				\EndIf
				\EndFor
				\State{Return $\pi$}
        \EndProcedure
        \Procedure{Sheepdog Update}{$\mathcal{G},x_{1}$}
        		\State{$v_{d_{1}}[k] \leftarrow$ attractive force from sheep~$i \in \mathcal{G}$ which is the farthest from $x_{1}$}
        		\State{$v_{d_{2}}[k] \leftarrow$ repulsive force from sheep which is the closest from sheepdog}
        		\State{$v_{d_{3}}[k] \leftarrow$ repulsive force from $x_{1}$}
        		\State{$v_{d}[k] \leftarrow K_{d_{1}}v_{1}[k] +K_{d_{2}}v_{2}[k] + K_{d_{3}}v_{3}[k]$}
        		\State{$x_{d}[k] \leftarrow x_{d}[k] + v_{d}[k]$}
        \EndProcedure
    \end{algorithmic}
    \end{algorithm}
    

Let $\mathcal{U} = \{1, \dotsc, N\}$ be the set of all sheep subscripts.
Let $\mathcal{I}[k]$ be the set of provisionally induced finished sheep at times $k$ and let $\mathcal{I}[0]= \emptyset$.
First, sheep with an index $\pi_{1}$ in the finite sequence~$\pi$ are targeted for tracking and the sheepdog moves to the location of the sheep using FAT. When the position of the sheepdog is less than a certain distance, i.e., if 
\begin{align}
\| x_{\pi_{1}}[k]-x_{d}[k]\| \leq r_{d}, \label{first_attack}
\end{align} 
then we determine that the tracking target sheep are provisionally guided, and add the sheep index to set~$\mathcal{I}[k]$, that is, update set~$\mathcal{I}$ as 
\begin{align}
\mathcal{I}[k + 1] = \mathcal{I}[k] \cup \{  \pi_{1}   \}.
\end{align}
Then, in the $\nu$th operation of Operation~A, the location of the sheep in $\pi_{\nu}$
is set as the provisional destination, and the sheep that have been already provisionally collected are guided to this provisional destination. The sheep-collection operation uses the farthest agent targeting method described in Section~\ref{tsunoda_method}.
If the positions of the provisional destination and all the tentatively guided sheep are less than a certain distance, i.e., if 
\begin{align}
\| x_{i}[k]-x_{\pi_{\nu}}[k]\| \leq g_{r} , i \in \mathcal{I}[k], 
\end{align}
 then the sheep that have been used as the provisional destination are considered to be provisionally guided sheep and added to the set $\mathcal I[k]$ as 
\begin{align}
\mathcal{I}[k + 1] = \mathcal{I}[k] \cup \{  \pi_{\nu}   \} . 
\end{align}
Finally, when all the sheep have been provisionally guided, that is, when
\begin{align}
\mathcal{U} = \mathcal{I}[k]
\end{align}
is reached, all the sheep are guided to their true destination position~$x_g$.
We remark that the operation of gathering the other sheep at the temporary destination $\pi_{\nu}$ is performed using the formulas \eqref{dog_x}, \eqref{v_d}, where at time~$k$ indices $\tilde{t}[k],\tilde{n}[k]$ of the furthest and latest sheep followed by the sheepdog are selected from the sheep in set~$\mathcal{I}[k]$ by
\begin{align}
\tilde{t}[k]  &= \argmax_{i \in \mathcal{I}[k]} \| x_{\pi_{\nu}}[k] - x_{i}[k]\|, \\
\tilde{n}[k]  &= \argmin_{i \in \mathcal{I}[k]} \| x_{d}[k] - x_{i}[k]\|, 
\end{align}
respectively.
In Operation~A, the velocity vector~$v_{d_{3}}[k]$ in Equation \eqref{dog3} is changed to 
\begin{align}
v_{d_{3}}[k] = \frac{x_{d}[k] - x_{\pi_{\nu}}[k]}{\| x_{d}[k] - x_{\pi_{\nu}}[k] \|}
\end{align}
to lead the other sheep to a temporary destination.
At the end of Operation~A, proceed to Operation~B and lead all the sheep to their original destination.
Algorithm~\ref{flowchart} shows the pseudocode illustrating the guidance strategy proposed in this study.

\section{Evaluation}

In this simulation, we compared the success rate of the guidance of a flock and the total distance traveled by the sheepdog to complete the guidance. Section~\ref{simlation_condition} describes the simulation conditions, and Section~\ref{simulation_result} presents the simulation results.

\subsection{Simulation conditions}\label{simlation_condition}

\begin{table}[t]
\centering
 \caption{Parameter configurations on simulation}
 \label{simu_para}
  \begin{tabular}{c|c|l}
   Variable           & Details                                                                    & Value           \\ \hline
   $N$             & Number of sheep                                                                   & $10, ~20, ~30, ~40,~50$        \\
   $\rho$         & Initial density to place the sheep flock($\times 10^{-4}$)      &  $6.0,~8.0,~10, ~12, ~14$ \\
   $x_{g}$          & destination                                                              & $[0,0]^{\top}$     \\
   $g_{r}$          & Destination radius                                                     & $20$        \\
   $r_{d}$          & Radius in equation \eqref{first_attack}                            & $30$        \\
   $T$             & Simulation Censoring Time                                        & $10000$   \\
   $x_{d}[0]$     & Initial coordinates for sheepdog                                  & $[-30,50]^{\top}$ \\
   $r_{s}$         & Radius at which sheep interact with each other             & $20$         \\
   $K_{s_{1}}$    & Repulsion gain of $v_{i}^{1}[k]$ in equation \eqref{eq:sheep_velo}       & $100$       \\
   $K_{s_{2}}$    & Alignment force gain of $v_{i}^{2}[k]$ in equation \eqref{eq:sheep_velo} & $0.5$        \\
   $K_{s_{3}}$    & Attractive gain of $v_{i}^{3}[k]$ in equation \eqref{eq:sheep_velo}       & $2.0$        \\
   $K_{s_{4}}$    & Repulsion gain of $v_{i}^{4}[k]$ in equation \eqref{eq:sheep_velo}       & $500$       \\
   $K_{d_{1}}$    & Attractive gain of $v_{d_{1}}[k]$ in equation \eqref{v_d}                   & $10$        \\
   $K_{d_{2}}$    & Repulsion gain of $v_{d_{2}}[k]$ in equation \eqref{v_d}                    & $1000$     \\
   $K_{d_{3}}$    & Repulsion gain of $v_{d_{3}}[k]$ in equation \eqref{v_d}                     & $4.5$        \\
  \end{tabular}
\end{table}

We compare the success rate of the guidance of a flock and the total distance traveled by the sheepdog to complete the guidance using the variables listed in  Table~\ref{simu_para}. 
Within the simulations, groups of sheep were randomly placed on a circle with radius $R_{{\rm init}} > 0$ and center $x_{g}$ according to a uniform distribution on the disk.
Based on the literature~\cite{Himo}, radius~$R_{{\rm init}}$ is determined by the following equation by using the number of sheep~$N$ and the initial density~$\rho$ to place the flock of sheep as 
\begin{align}
R_{{\rm init}} =\sqrt{\frac{N}{\pi\rho}}. \label{eq:radius}
\end{align}
The number~$N$ of sheep and the initial density~$\rho$ at which the flock is placed are the parameters expressing the ease of guidance. The higher the density of the sheep flock at the initial location, the denser the flock, and thus, the easier the guidance. The larger $N$, the more complex the expected guidance.

Five different values of~$N$ and~$\rho$ were prepared, as listed in Table~\ref{simu_para}, and the total of $5\times 5$ combinations of these parameters were used in the guidance control to evaluate the performance.
Table~\ref{tb:R} lists $R_{{\rm init}}$ obtained using Equation \eqref{eq:radius} when $N$ and $\rho$ are used.

\begin{table}[t]
  \centering
  \caption{Circle radius $R_{{\rm init}}$ for density~$\rho$ and number of sheep}
  \vspace{3mm}
  \begin{tabular}{|c|c|c|c|c|c|} \hline
    \diagbox{$\rho$}{$N$} & 10 & 20 & 30 & 40 & 50 \\ \hline
    0.00060 & 72.84   & 103.0   & 126.2 & 145.7& 162.9\\ \hline
    0.00080 & 63.08   & 89.21   & 109.3 & 126.2& 141.0\\ \hline
    0.00100 & 56.42   & 79.79   & 97.72 & 112.8& 126.2\\ \hline
    0.00120 & 51.50   & 72.84   & 89.21 & 103.0& 115.2\\ \hline
    0.00140 & 47.68   & 67.43   & 82.59 & 95.47& 106.6\\ \hline
  \end{tabular}
  \label{tb:R}
\end{table}

For a flock of sheep artificially placed using the above method, the position was updated in 50 steps using the control law in Equation \eqref{eq:position} to achieve a more realistic initial placement.
Then, Reverse, Exchange, and Jump were used as the array manipulation algorithm in the RLS method, and the number of updates was set to 10,000 times. The proposed algorithm was then used to perform guidance control.

The method to be compared is the original FAT method~\cite{Tsunoda_near_kai}.
The proposed and conventional methods were tested 100 times each, and the success rate of guidance and the total distance traveled by the sheepdog were evaluated.
Successful guidance was determined based on whether the guidance is completed within a pre-specified time~$T$. Equation \eqref{dog_JK} evaluates the total distance traveled by the sheepdog.

\begin{figure}[tb]
\centering
\includegraphics[clip,scale=0.43]{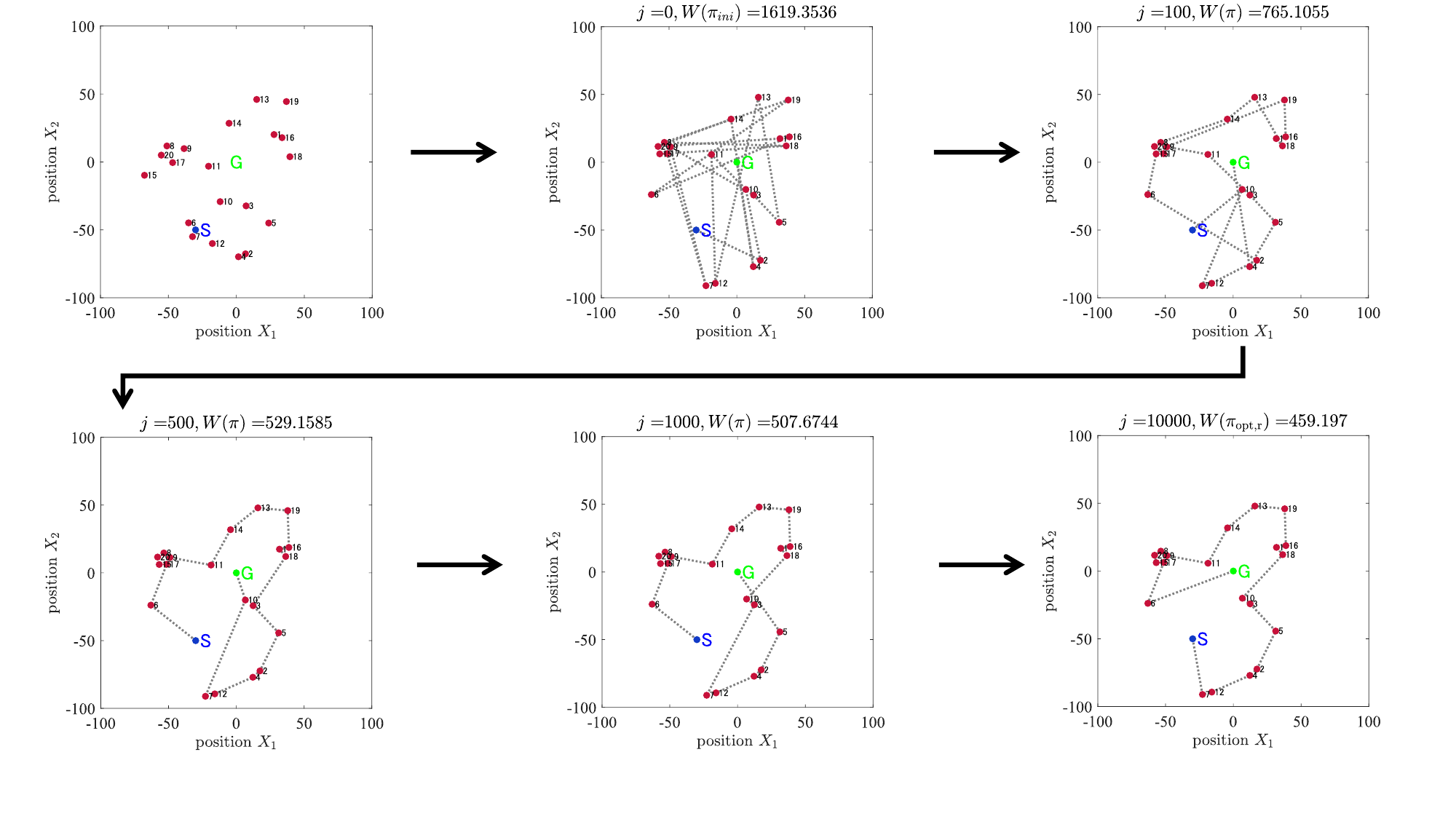}
\caption{Result of designing guidance route using Reverse, blue circle; position of sheepdog, red circles; position of sheep, S; start point of the sheepdog, G; goal, dash line; a guidance route}
\label{sinir}
\end{figure}

\begin{figure}[tb]
\centering
\includegraphics[clip,scale=0.43]{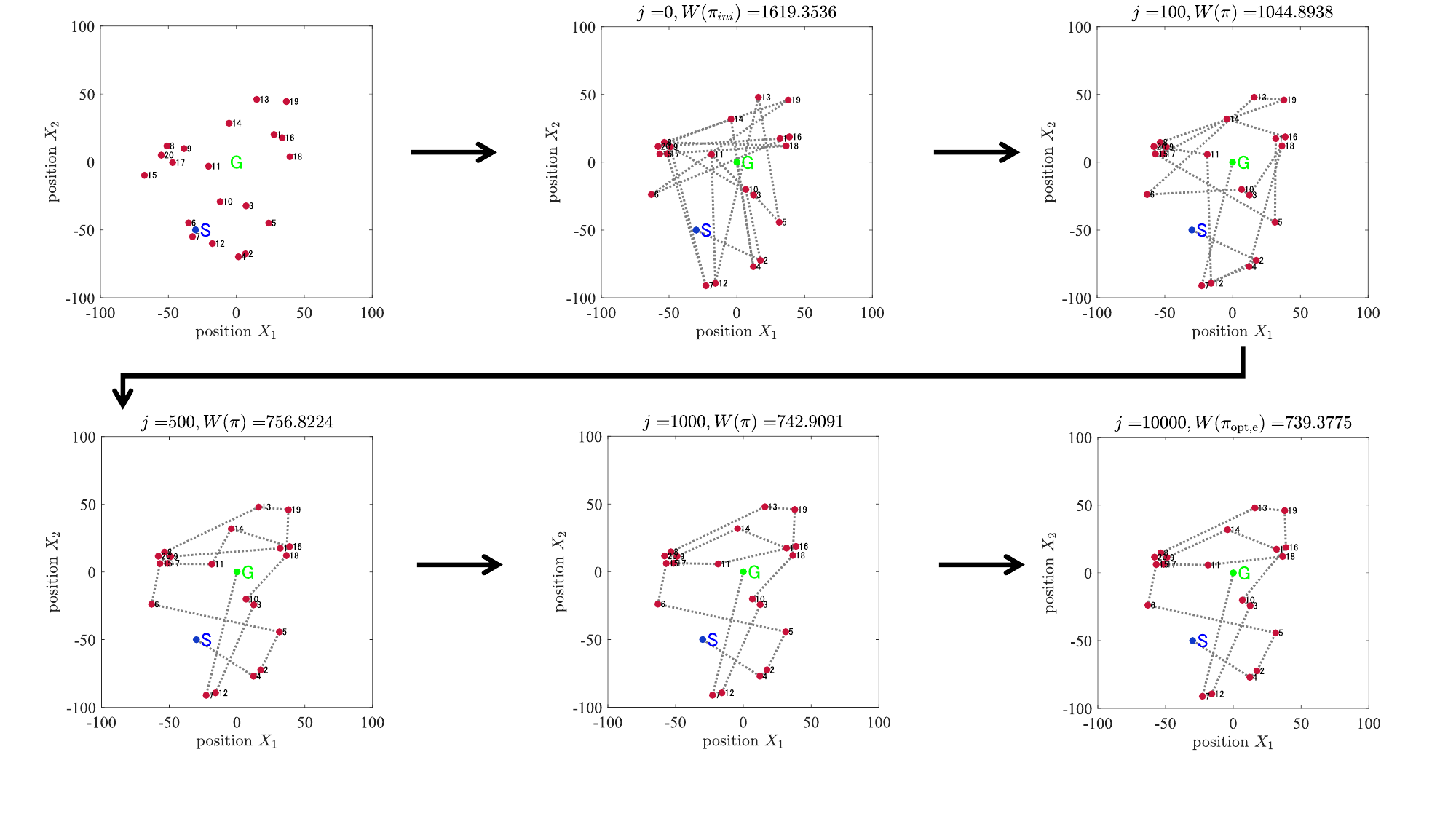}
\caption{Result of designing guidance route using Exchange, blue circle; position of sheepdog, red circles; position of sheep, S; start point of the sheepdog, G; goal, dash line; a guidance route}
\label{sheep_ini_TSP_exchange}
\end{figure}

\begin{figure}[tb]
\centering
\includegraphics[clip,scale=0.43]{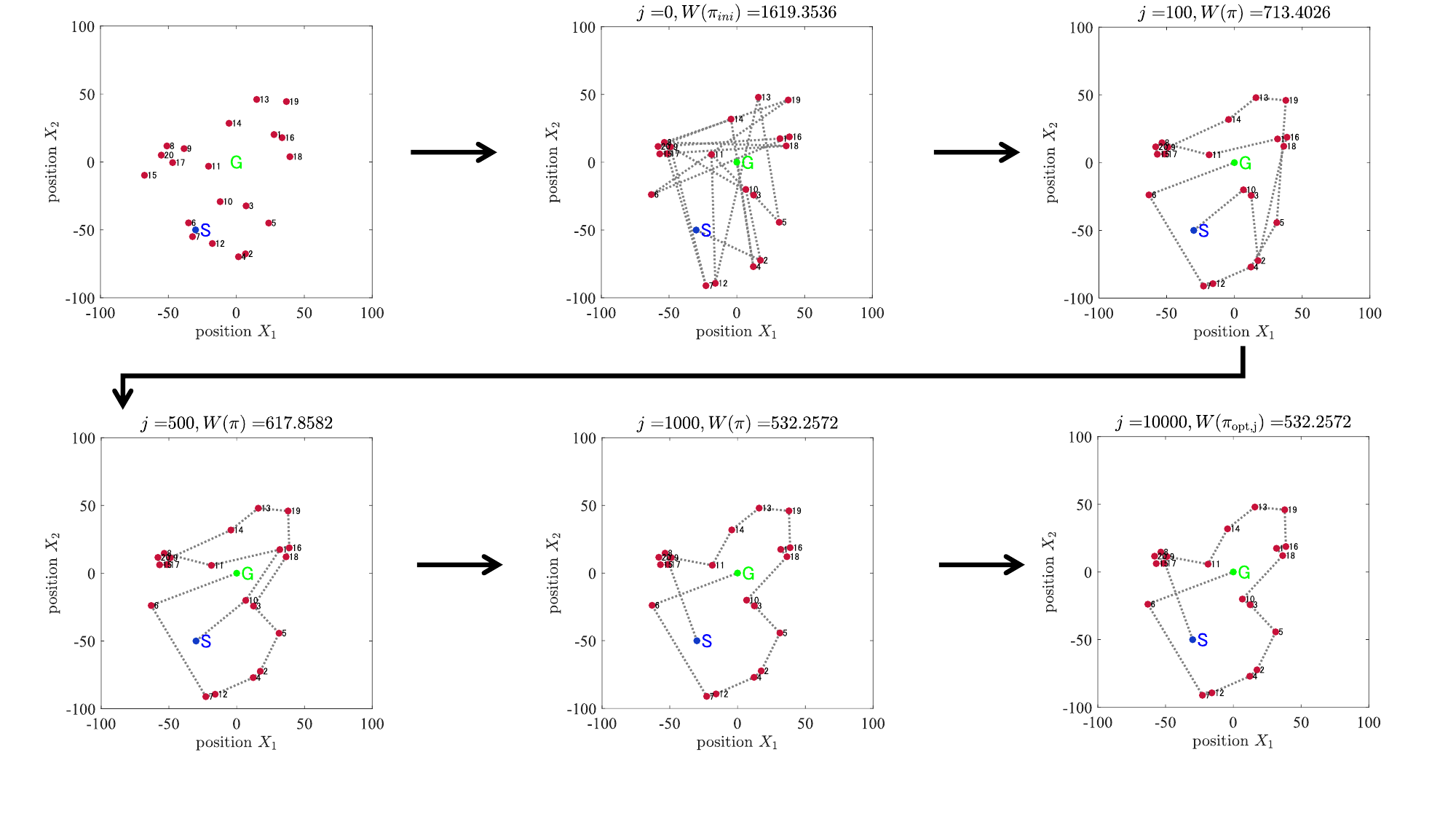}
\caption{Result of designing guidance route using jump, blue circle; position of sheepdog, red circles; position of sheep, S; start point of the sheepdog, G; goal, dash line; a guidance route}
\label{sheep_ini_TSP_jump}
\end{figure}


\begin{figure}[tb]
  \begin{minipage}[b]{0.33\linewidth}
 \centering
  \includegraphics[clip,scale=0.33]{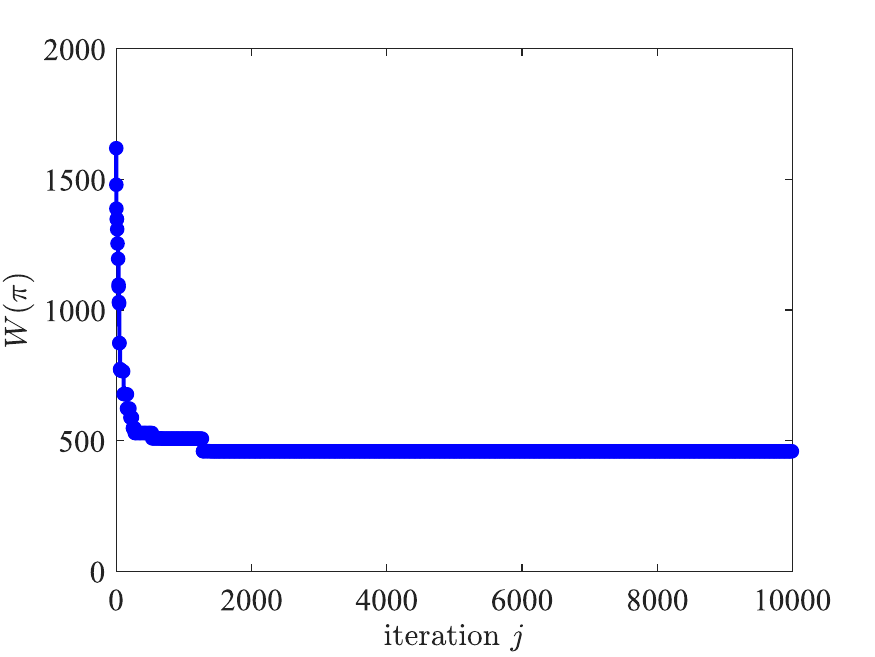}
  Reverse\\
    \end{minipage}
  \begin{minipage}[b]{0.33\linewidth}
 \centering
  \includegraphics[clip,scale=0.33]{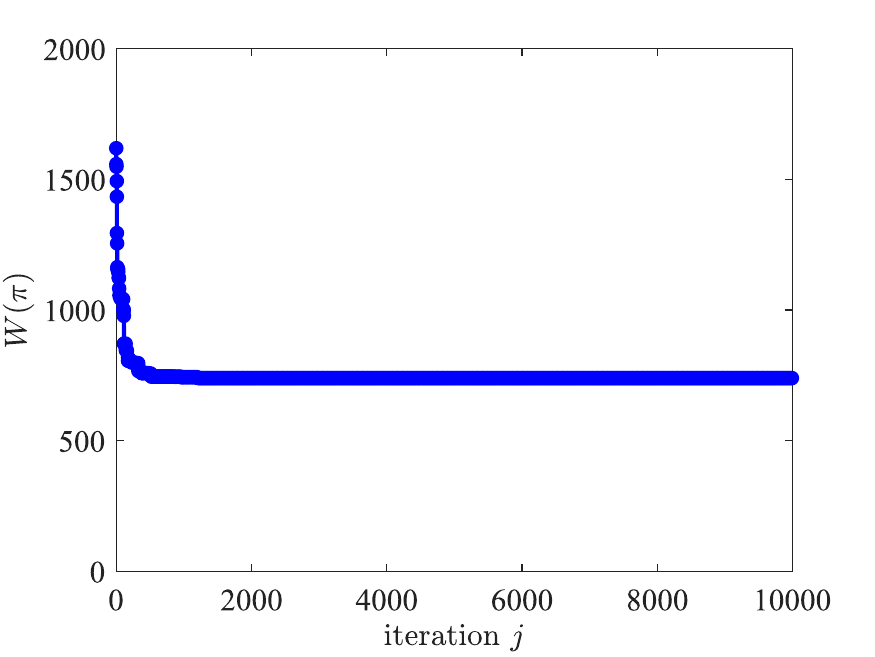}
  Exchange\\
    \end{minipage}
  \begin{minipage}[b]{0.33\linewidth}
 \centering
  \includegraphics[clip,scale=0.33]{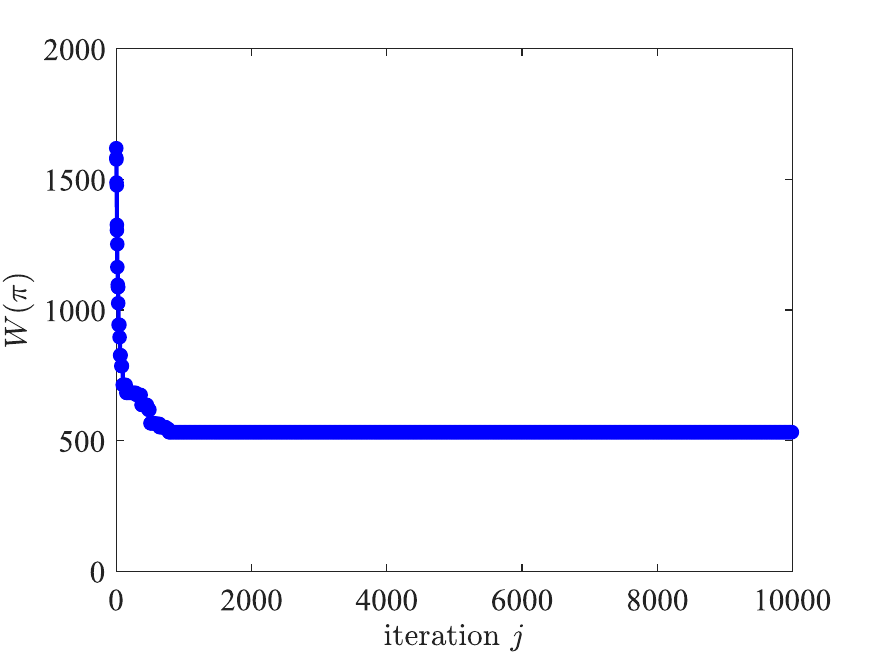}
  Jump\\
    \end{minipage}
    \caption{Result of cost graph (left: Reverse, center: Exchange, right: Jump)}
    \label{cost_TSP_real}
\end{figure}

\subsection{Simulation Results}\label{simulation_result}
First, the operation of the proposed method was verified.
Here, $N=20$, $\rho=0.00120$, and $R_{{\rm init}}=72.84$, and the sheep flocks were randomly arranged.
Figures~\ref{sinir}, \ref{sheep_ini_TSP_exchange}, and~\ref{sheep_ini_TSP_jump} show the optimization of the guided path for the initial placement of the sheepdog and sheep.
The upper left of Figures~\ref{sinir}, \ref{sheep_ini_TSP_exchange}, and~\ref{sheep_ini_TSP_jump} 
show the results of the artificial placement of the sheepdog and sheep describe in Section~\ref{simlation_condition}.
The position of the sheep was then updated for 50 steps based on the sheep control law described in~\eqref{eq:position}.
Consequently, the position of the sheep was updated to the position in the upper center graph of Figures~\ref{sinir}, \ref{sheep_ini_TSP_exchange}, and~\ref{sheep_ini_TSP_jump}.
Then, from the positions of the sheepdog and sheep, the finite sequence~$\pi_{{\rm ini}}$ and guidance cost $W(\pi_{{\rm ini}})$ in the initial guidance path of the Traveling Salesman Problem are obtained as follows:
\begin{align}
\pi_{{\rm ini}}    &=[2,10,11,12,13,5,3,9,7,17,15,18,8,19,6,16,1,20,14,4],  \nonumber \\
W(\pi_{{\rm ini}})&=1619.3436. 
\end{align}
Subsequently, based on the proposed method, the finite sequence was updated using RLS to reduce the guidance cost.
Figures~\ref{sinir}, \ref{sheep_ini_TSP_exchange}, and~\ref{sheep_ini_TSP_jump} from top right to bottom right show the optimization.
$i$ on each graph indicates the number of times the optimization is updated.
The guided path changed as the optimization progressed.
Finally, Figures~\ref{sinir}, \ref{sheep_ini_TSP_exchange}, and~\ref{sheep_ini_TSP_jump} show the lower right corner and finite sequence~$\pi_{{\rm opt,r}}$, $\pi_{{\rm opt,e}}$, $\pi_{{\rm opt,j}}$ and the guidance cost 
$W(\pi_{{\rm opt,r}})$, $W(\pi_{{\rm opt,e}})$, $W(\pi_{{\rm opt,j}})$ at that time were computed to be
\begin{align}
\pi_{{\rm opt,r}}    &=[7,12,4,2,5,3,10,18,1,16,19,13,14,11,9,8,20,15,17,6],  \nonumber  \\ 
W(\pi_{{\rm opt,r}})&=459.1970, \label{reverse}\\
\pi_{{\rm opt,e}}    &=[4, 2, 5, 6, 15, 17, 11, 18, 1, 14,9, 20, 8, 13, 19, 16, 10, 3, 12, 7],  \nonumber  \\ 
W(\pi_{{\rm opt,e}})&=739.3775, \label{exchange}\\
\pi_{{\rm opt,j}}    &=[8, 20, 15, 17, 9, 11, 14, 13, 19, 16,1, 18, 10, 3, 5, 2, 4, 12, 7, 6],  \nonumber \\ 
W(\pi_{{\rm opt,j}})&=532.2572,\label{jump}
\end{align}
respectively.
Figure~\ref{cost_TSP_real} shows how guidance cost $W(\pi)$ improved as we repeat the reordering process. We observe that all guidance costs $W(\pi)$ converge after approximately $1,000$ iterations.

\begin{figure}[tb]
 \centering
  \includegraphics[clip,scale=0.43]{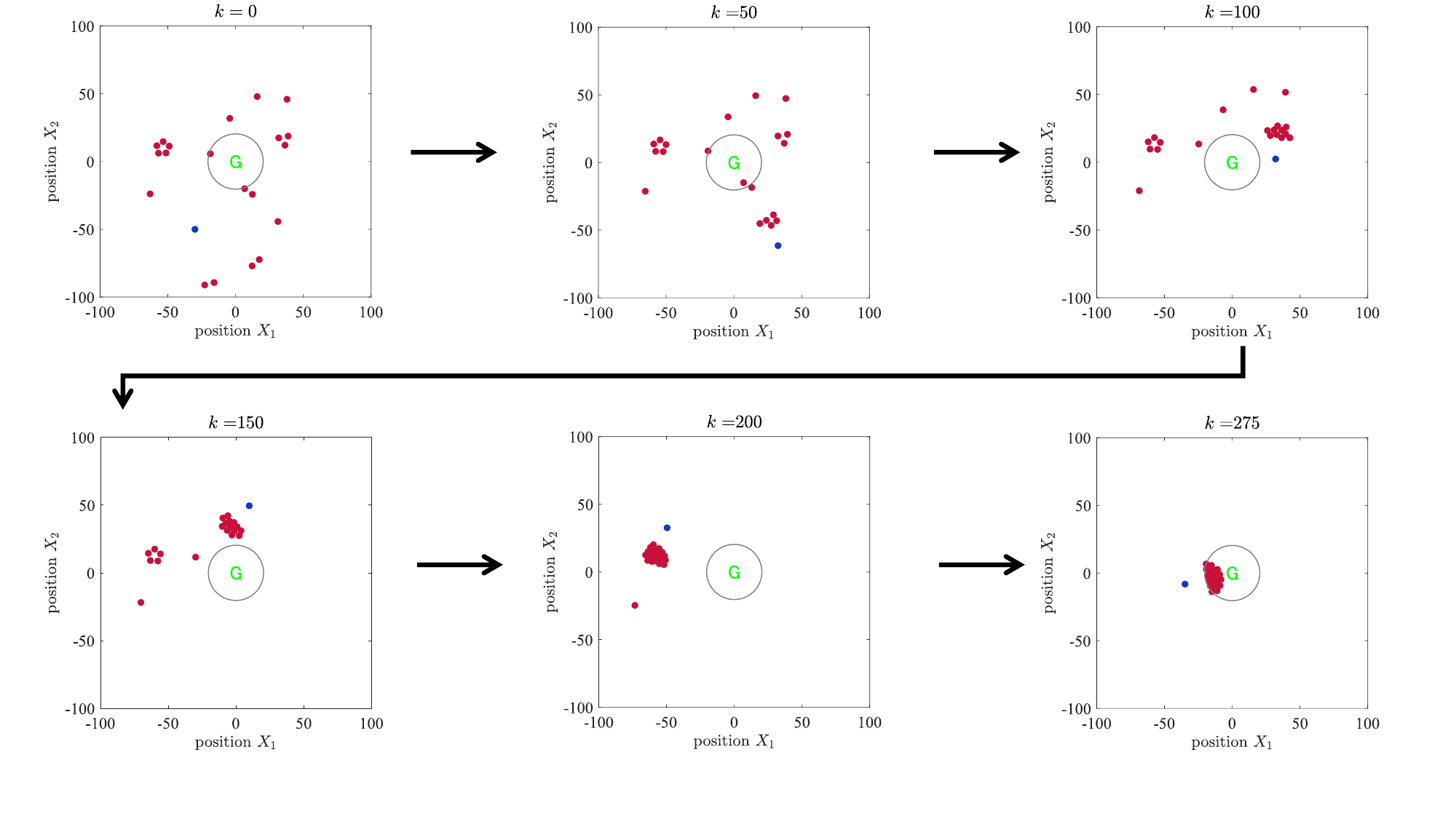}
  \caption{Result of guidance control by the proposed method reverse, blue circle; position of sheepdog, red circles; position of sheep, G; goal}
  \label{TSP_result_reverse}
\end{figure}

\begin{figure}[tb]
 \centering
  \includegraphics[clip,scale=0.43]{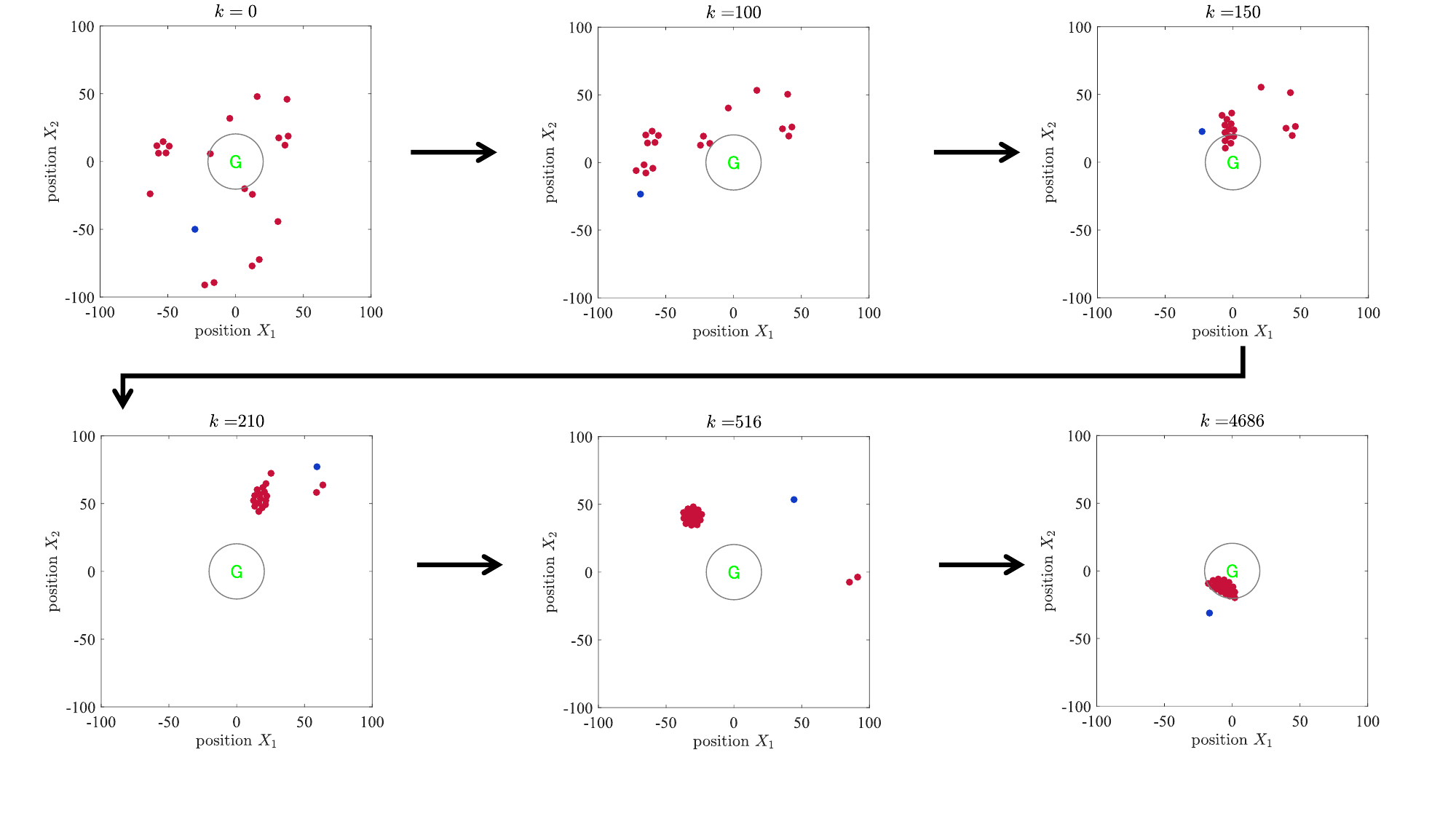}
  \caption{Result of guidance control using the proposed method exchange, blue circle; position of sheepdog, red circles; position of sheep, G; goal} 
  \label{TSP_result_exchange}
\end{figure}

\begin{figure}[tb]
 \centering
  \includegraphics[clip,scale=0.43]{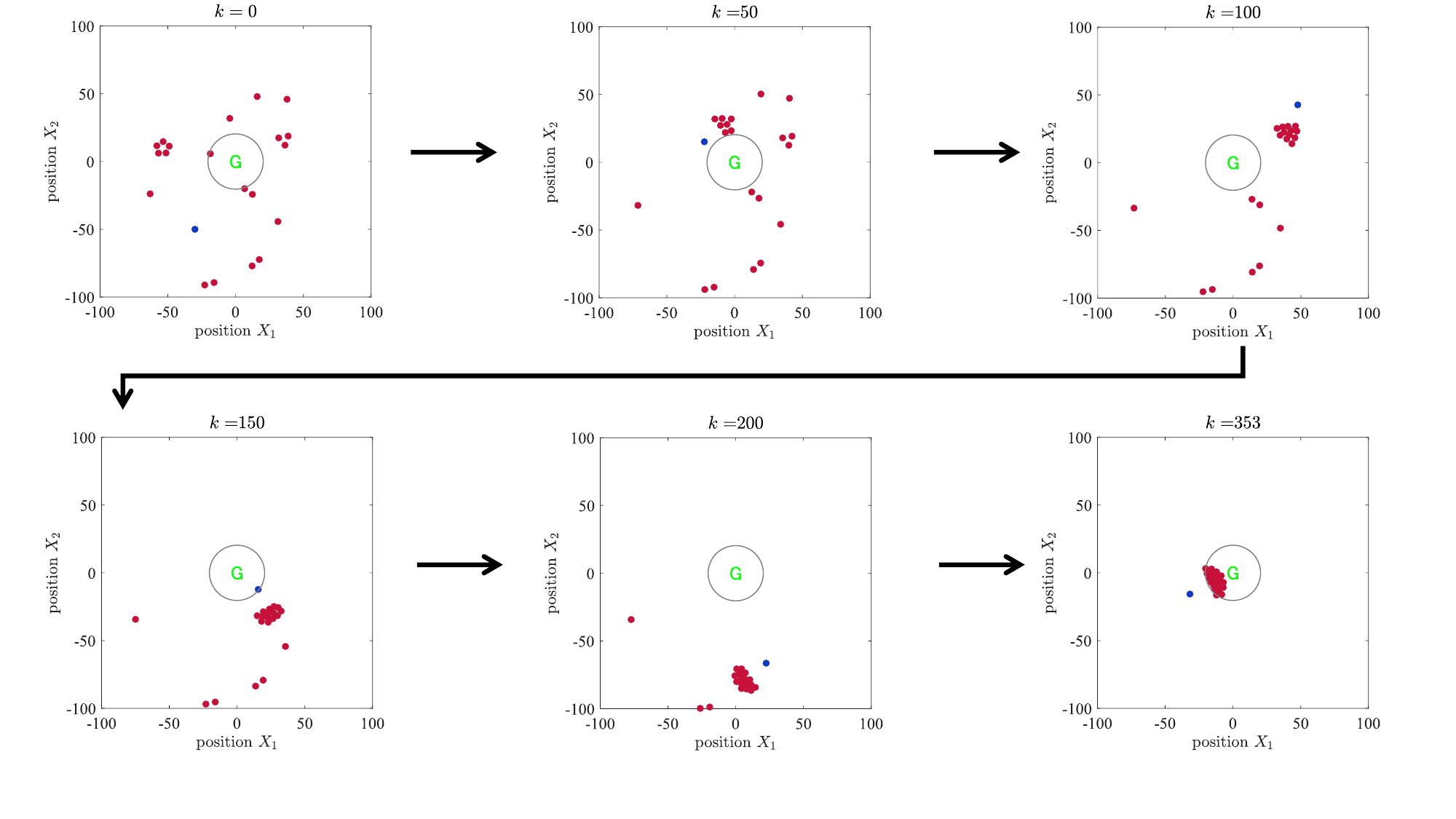}
  \caption{Result of guidance control using the proposed method jump, blue circle; position of sheepdog, red circles; position of sheep, G; goal}
  \label{TSP_result_jump}
\end{figure}

\begin{figure}[tb]
 \centering
  \includegraphics[clip,scale=0.43]{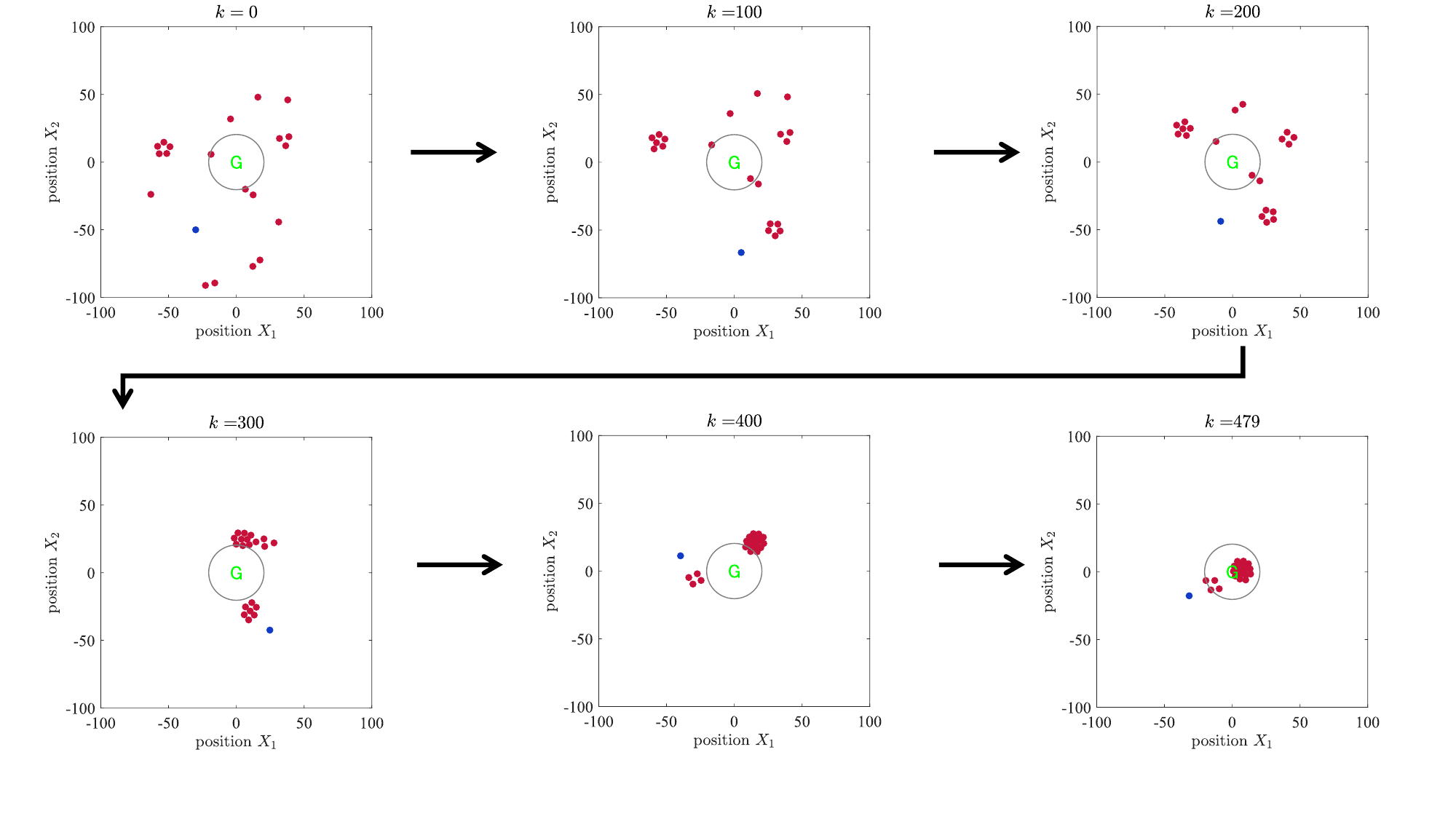}
  \caption{Result of guidance control using FAT, blue circle; position of sheepdog, red circles; position of sheep, G; goal}
  \label{FAT_result}
\end{figure}

\begin{figure}[tb]
  \begin{minipage}[b]{0.45\linewidth}
 \centering
  \includegraphics[clip,scale=0.5]{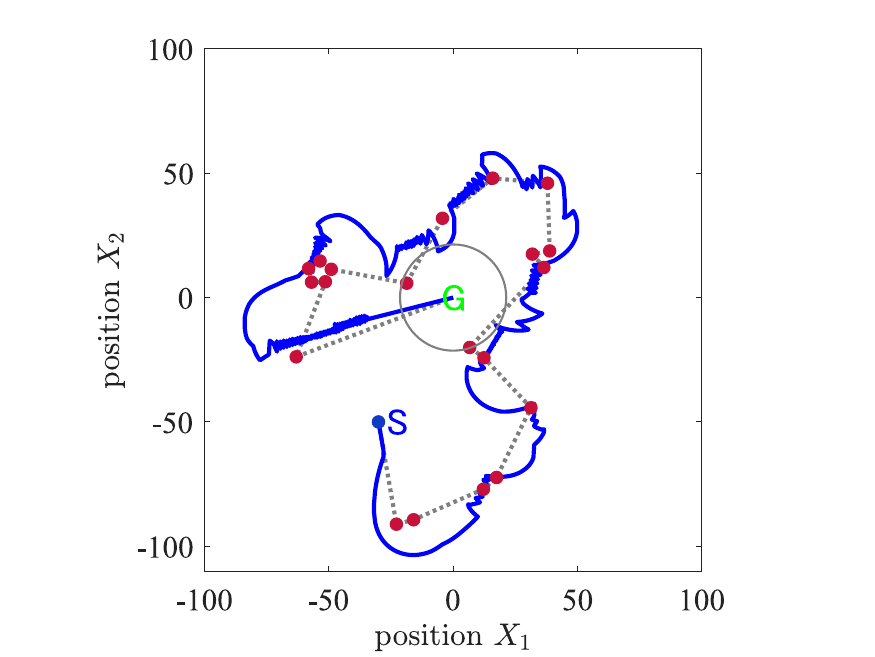}
\hspace{1cm} proposed method (reverse)\hspace{-1cm}
  \end{minipage}
    \begin{minipage}[b]{0.45\linewidth}
 \centering
  \includegraphics[clip,scale=0.5]{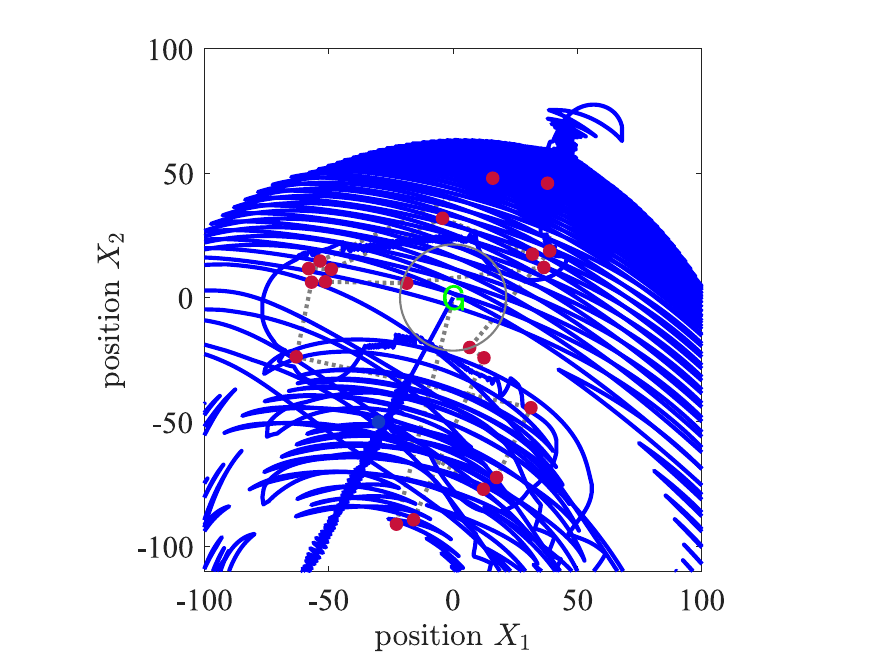}
 \hspace{1cm}proposed method (exchange)\hspace{-1cm}
  \end{minipage}\\
  \begin{minipage}[b]{0.45\linewidth}
 \centering
  \includegraphics[clip,scale=0.5]{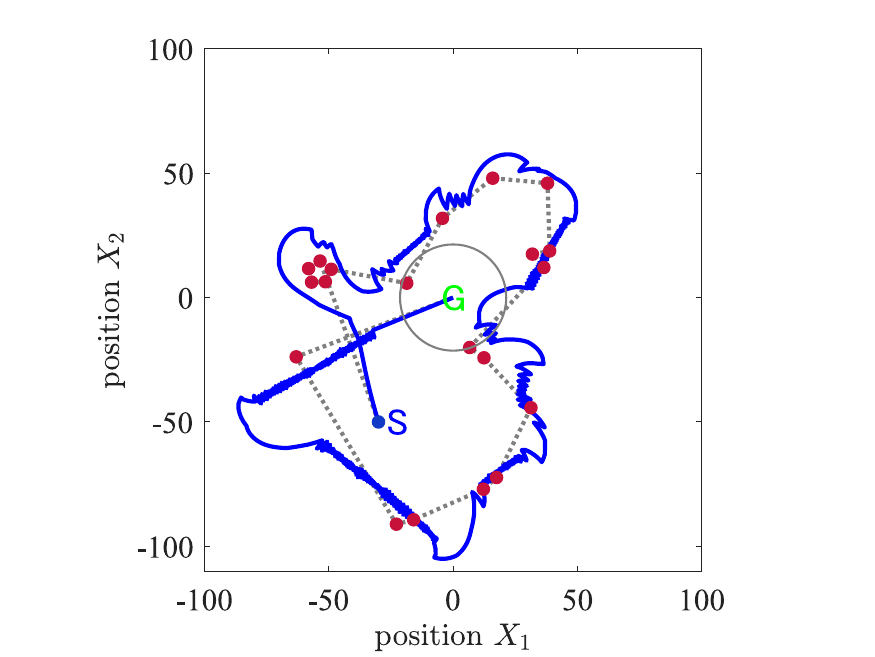}
      \hspace{1cm}proposed method (jump)\hspace{-1cm}
  \end{minipage}
    \begin{minipage}[b]{0.45\linewidth}
 \centering
  \includegraphics[clip,scale=0.5]{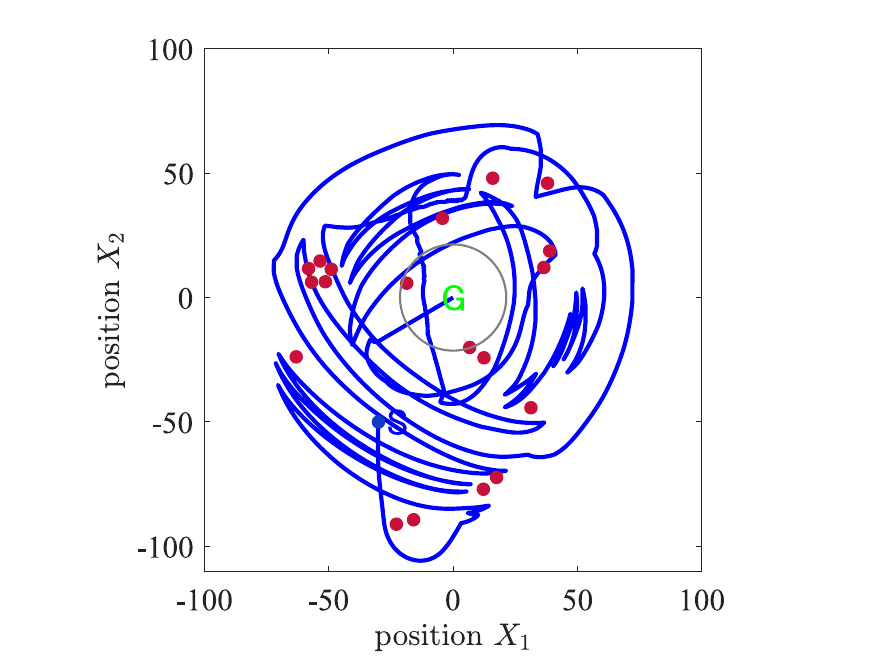}
 \hspace{1cm}conventional method (FAT)\hspace{-1cm}
  \end{minipage}
    \caption{Trajectory of sheepdog, blue circle; position of sheepdog, red circles; position of sheep, S; start point, 
    G; goal, blue line; real guidance route, dash line; desire guidance route}
    \label{trajectory}
\vspace{2mm}

  \begin{minipage}[b]{0.33\linewidth}
 \centering
  \includegraphics[clip,scale=0.3]{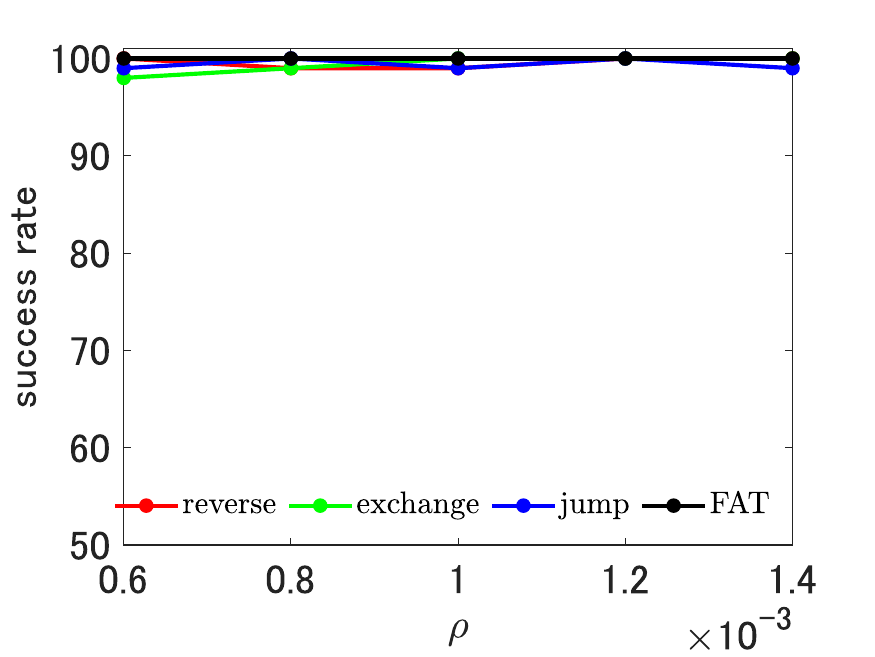}
 {\small $N=10$}\\
    \end{minipage}
    \begin{minipage}[b]{0.33\linewidth}
 \centering
  \includegraphics[clip,scale=0.3]{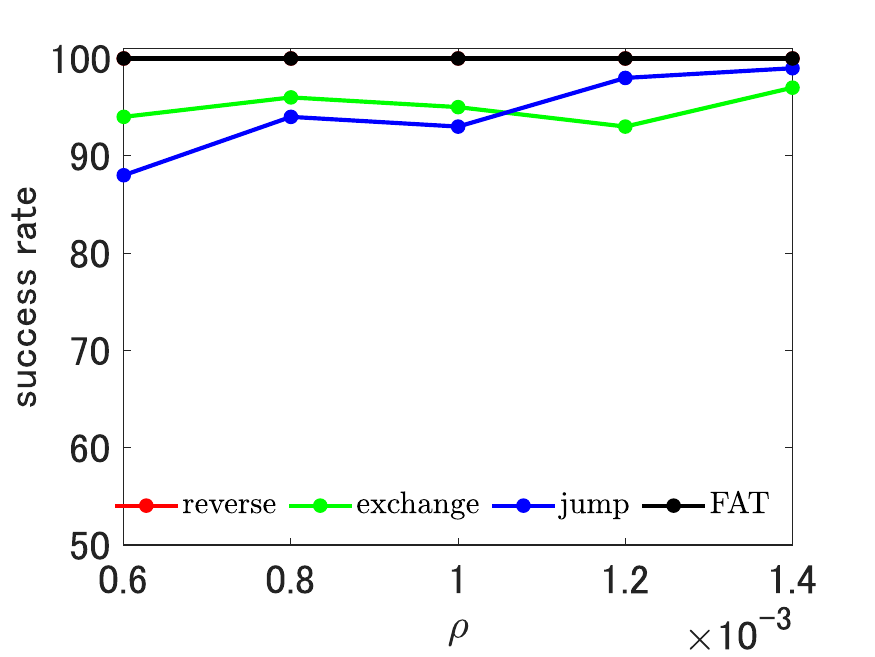}
 {\small $N=20$}\\
    \end{minipage}
        \begin{minipage}[b]{0.33\linewidth}
 \centering
  \includegraphics[clip,scale=0.3]{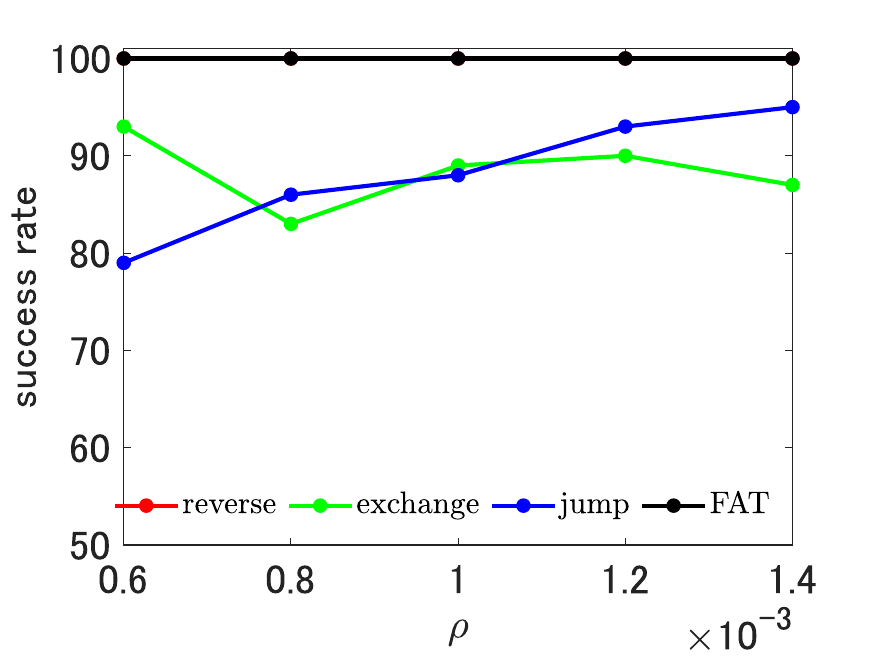}
 {\small $N=30$}\\
    \end{minipage}\\
      \begin{minipage}[b]{0.5\linewidth}
 \centering
  \includegraphics[clip,scale=0.3]{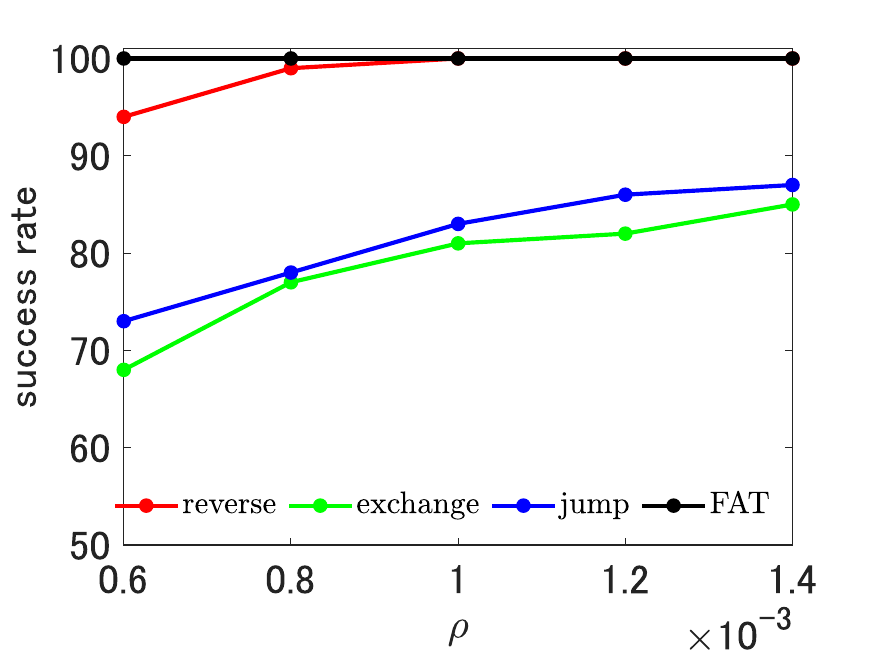}\\
 {\small $N=40$}\\
    \end{minipage}
    \begin{minipage}[b]{0.5\linewidth}
 \centering
  \includegraphics[clip,scale=0.3]{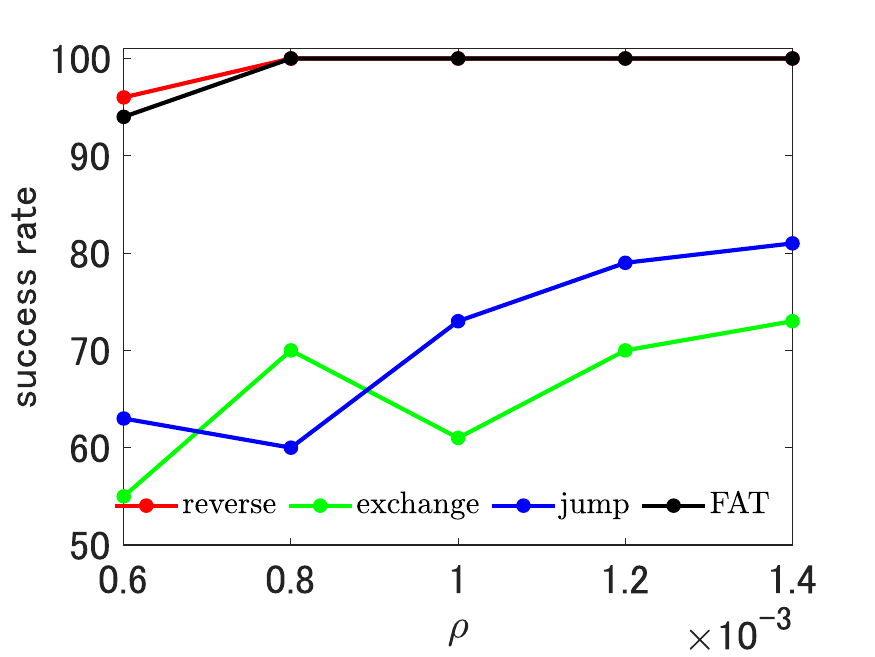}\\
 {\small $N=50$}\\
    \end{minipage}
  \caption{Results showing guidance success rate under various conditions}
  \label{success}
  \end{figure}

The results of the induced control of the proposed method using a finite sequence of Equations \eqref{reverse}, \eqref{exchange}, and \eqref{jump} are shown in Figures~\ref{TSP_result_reverse}, \ref{TSP_result_exchange}, and~\ref{TSP_result_jump}, respectively. In the figures, 
the guidance control is shown along the arrow starting from the top left. 
We observe that, in the proposed method, all sheep entered within a radius of $20$ from the destination~$x_{g}$, and guidance was completed. The Reverse method (Figure~\ref{TSP_result_reverse}) took approximately $k=275$ steps, the Exchange method (Figure~\ref{TSP_result_exchange}) incurred $k=4686$ steps, and the Jump method (Figure~\ref{TSP_result_jump}) incurred $k=353$ steps.
Figure~\ref{FAT_result} shows the results of the inductive control using the FAT method, which is a comparative method.
Inductive control was completed at $k=479$.

Figure~\ref{trajectory} shows the actual guided path taken by the sheepdog when the proposed and FAT methods were used.
Although the FAT method targets the sheep farthest from the destination and gradually leads them to the destination, which therefore tends a long distance traveled by the sheepdog, the proposed method allows the sheepdog to achieve guidance with a significantly smaller traveling distance.

\begin{figure}[tb]

    \begin{minipage}[b]{0.33\linewidth}
 \centering
  \includegraphics[clip,scale=0.3]{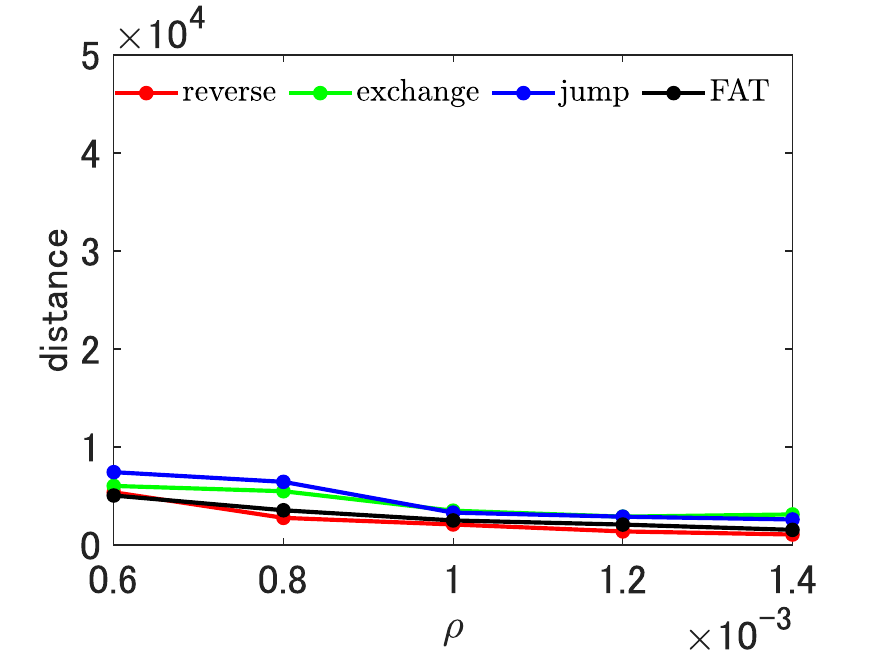}
 $N=10$\\
    \end{minipage}
    \begin{minipage}[b]{0.33\linewidth}
 \centering
  \includegraphics[clip,scale=0.3]{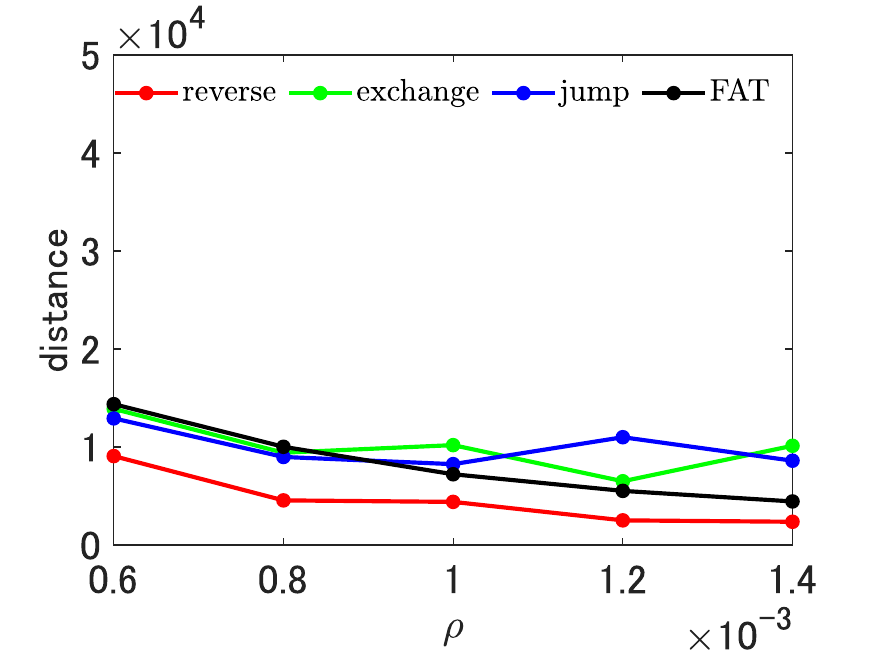}
 $N=20$\\
    \end{minipage}
        \begin{minipage}[b]{0.33\linewidth}
 \centering
  \includegraphics[clip,scale=0.3]{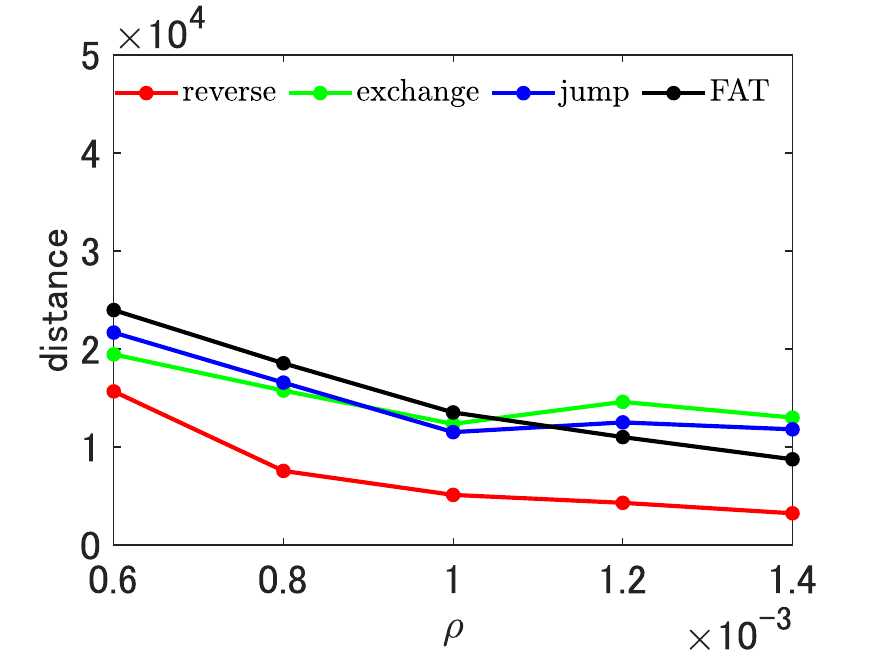}
 $N=30$\\
    \end{minipage}\\
      \begin{minipage}[b]{0.5\linewidth}
 \centering
  \includegraphics[clip,scale=0.3]{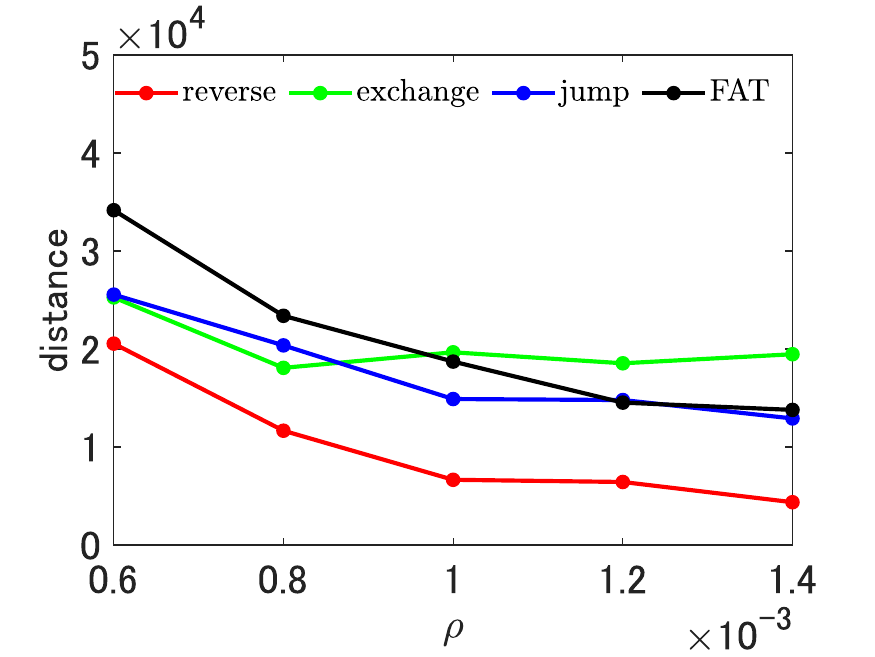}\\
 $N=40$\\
    \end{minipage}
    \begin{minipage}[b]{0.5\linewidth}
 \centering
  \includegraphics[clip,scale=0.3]{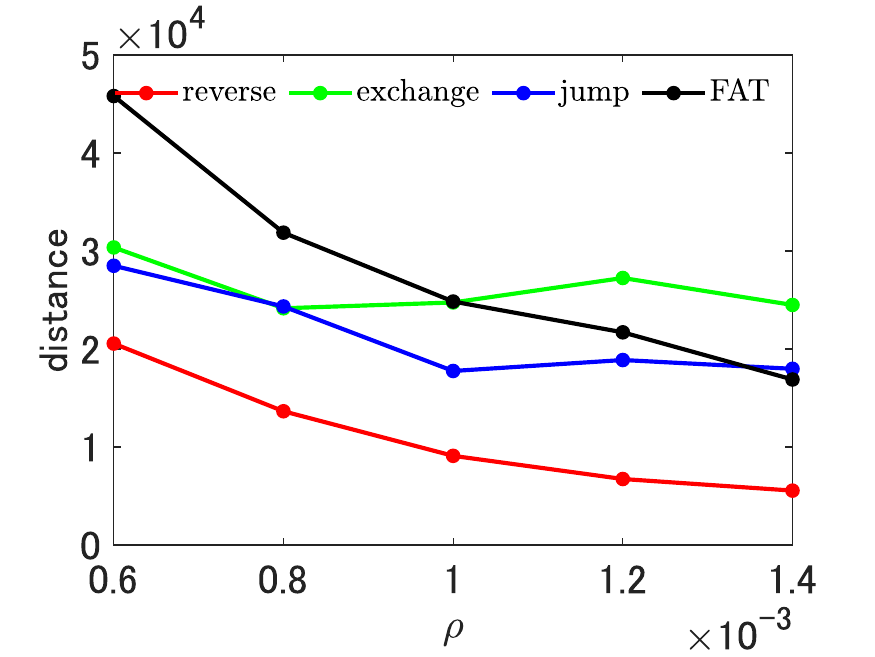}\\
 $N=50$\\
    \end{minipage}
        \caption{Results showing the total distance by sheepdog under various conditions
        \label{totaldistance}
        }
\end{figure}

To further evaluate the effectiveness of the proposed method, we perform a through evaluation, in which we perform 100 simulations for each of the parameters by varying the initial placements of the sheep and sheepdog. The results are shown in Figures~\ref{success} and~\ref{totaldistance}.
In both figures, the horizontal axis represents the initial sheep density~$\rho$ and the vertical axis represents the number of sheep~$N$.
Figure~\ref{success} shows the success rate of shepherding and Figure~\ref{totaldistance} shows the average total distance traveled by a sheepdog.
The success rate of shepherding is high, almost 100$\%$, for both the proposed and the comparison methods. However, the total distance traveled by the sheepdog is almost the same for the proposed and the comparison methods when the number of sheep was $N=10$; however, as $N$ increases, the total distance traveled by the sheepdog is smaller for the proposed method than for the comparison methods, regardless of the value of $\rho$.
Therefore, the effectiveness of the proposed method was confirmed through this simulation.

\section{Conclusion}
In this study, we addressed the problem of designing a guidance route for a sheepdog system, a multiagent system in which a sheepdog can reduce the traveling distance traversed to guide a sheep to its destination.

In the proposed method, the traveling distance required for a sheepdog to guide all the sheep was quantified using TSP, and a finite sequence to reduce the traveling distance was designed through evolutionary computation using the RLS method.
The travel distance could be reduced while maintaining a high guidance success rate by adopting a modified version of the guidance strategy of Himo et al. and using the FAT method for guidance control.

The effectiveness of the proposed method was confirmed by simulation, and it was found that the distance traveled by the sheepdog to guide the sheep was shorter with the proposed method than with the comparison method.

Future tasks include constructing a framework that allows the sheepdog to sequentially update the guidance paths for guiding the sheep while performing guidance control, rather than designing them before guidance control, and verifying the effectiveness of the proposed method through actual experiments using a mobile robot.

\section*{acknowledgment}
This work was partially supported by JSPS Grants-in-Aid for Scientific Research 22K14282, 21H01352, and 22K14277.


\begin{thebibliography}{99}
\bibitem{Notomista}
G. Notomista, S. Mayya, Y. Emam, C. Kroninger, A. Bohannon, S. Hutchinson, and M. Egerstedt, ``A resilient and energy-aware task allocation framework for heterogeneous multi-robot systems,'' {\it IEEE Transactions on Robotics}, Vol. 38, No. 1, pp. 159--179, 2022.
\bibitem{Sugano}
R. Sugano, N. Takeuchi, K. Sekiguchi, K. Nonaka, ``Layer structured multiple-UAV system mixing centralized and distributed agents,'' In {\it 2020 IEEE/SICE International Symposium on System Integration}, pp. 915--920, 2020.
\bibitem{Ikeda}
T. Ikeda, M. Nagahara, and K. Kashima, ``Maximum hands-off distributed control for consensus of multi-agent systems with sampled-data state observation,''
{\it IEEE Transactions on Control of Network Systems},  Vol. 6, No. 2, pp. 852--862, 2019.
\bibitem{Cai}
K. Cai, and M. Nagahara. ``A new perspective on cooperative control of multi-agent systems through different types of graph Laplacians,'' {\it Advanced Robotics}, Vol. 37.1--2, pp. 2--11, 2023.
\bibitem{Atman}
M. W. S. Atman, T. Hatanaka, Z. Qu, N. Chopra, J. Yamauchi, and M. Fujita, 
``Motion synchronization for semi-autonomous robotic swarm with a passivity-short human operator,''
{\it International Journal of Intelligent Robotics and Applications}, pp. 235--251, 2018.
\bibitem{Kyo}
K. K. Oh, M. C. Park, and H. S. Ahn, ``A survey of multi-agent formation control,'' {\it Automatica}, 
Vol. 53, pp. 424--440, 2015.
\bibitem{shepherding_survey}
N. K. Long, K. Sammut, D. Sgarioto, M. Garratt, H. A. Abbass, ``A comprehensive review of shepherding as a bio-inspired swarm-robotics guidance approach,'' {\it IEEE Transactions on Emerging Topics in Computational Intelligence}, Vol. 4, No. 4, pp. 523--537, 2020.
\bibitem{information sensor}
A. J. Hepworth, A. Hussein, D. J. Reid, and H. A. Abbass,``Swarm analytics: Designing information markers to characterise swarm systems in shepherding contexts,''
{\it Adaptive behavior}, Vol. 31, No. 4, pp. 323--349, 2023.
\bibitem{Shepherding_PSO}
R. E. Mohamed, R. Hunjet, S. Elsayed, H. A. Abbass,``
Connectivity-Aware particle swarm optimisation for swarm shepherding,''
{\it IEEE Transactions on Emerging Topics in Computational Intelligence}, Vol. 7, No. 3, pp. 661--683, 2023.
\bibitem{Shepherding_RL}
A. Hussein, E. Petraki, S. Elsawah, H. A. Abbass,``
Autonomous swarm shepherding using curriculum-based reinforcement learning,''
In {\it Proceedings of the 21st International Conference on Autonomous Agents and Multiagent Systems}, pp. 633--641, 2022.
\bibitem{strombom}
D. Strömbom, R. P. Mann, A. M. Wilson, S. Hailes, A. J. Morton, D. J. T. Sumpter, and A. J. King,
``Solving the shepherding problem: heuristics for herding autonomous interacting agents,'' {\it Journal of the Royal Society Interface}, Vol. 11, No. 100, 2014.
\bibitem{Tsunoda}
Y. Tsunoda, Y. Sueoka, Y. Sato, and K. Osuka, 
``Analysis of local-camera based shepherding Navigation,'' 
{\it Advanced Robotics}, Vol. 32. No. 23, pp. 1217--1228, 2018.
\bibitem{Tsunoda_near}
Y. Tsunoda, M. Ishitani, Y. Sueoka, and K. Osuka, ``Analysis of sheepdog-type navigation for a sheep model with dynamics,'' 
In {\it Twenty-Fourth International Symposium on Artificial Life and Robotics}, pp. 499--503, 2019.
\bibitem{Tsunoda_near_kai}
Y. Tsunoda, M. Ishitani, Y. Sueoka, and K. Osuka, ``Analysis of sheepdog-type navigation for minimal sheep model,'' In {\it 3rd International Symposium on Swarm Behavior and Bio-Inspired Robotics}, pp. 197--200, 2019. 
\bibitem{Fujioka}
A. Fujioka, M. Ogura, and N. Wakamiya, ``Shepherding algorithm for heterogeneous flock with model-based discrimination,'' 
{\it Advanced Robotics}, vol. 37, No. 1--2, pp. 99--114, 2023.
\bibitem{Himo}
R. Himo, M. Ogura, and N. Wakamiya, ``Iterative algorithm for shepherding unresponsive sheep,'' 
{\it Mathematical Biosciences and Engineering}, 
Vol. 19, No. 4, pp. 3509--3525, 2022.
\bibitem{Tsunoda_theoretical}
Y. Tsunoda, Y. Sueoka, T. Wada, K. Osuka, ``Theoretical analysis of mobile control method for group agents motivated by sheepdog shepherding,'' {\it Transactions of the Society of Instrument and Control Engineers}, Vol. 55, No. 8, pp. 507--515, 2019.
\bibitem{Flock_control}
R. Vaughan, N. Sumpter, A. Frost, and S. Cameron, ``Robot
sheepdog project achieves automatic flock control,'' In {\it Fifth International Conference on the Simulation of Adaptive Behaviour}, 1998.
\bibitem{Tsunoda_ex}
Y. Tsunoda, Y. Sueoka, T. Wada, K. Osuka, ``Sheepdog-type robot navigation: Experimental verification based on a linear model,'' In
{\it 2020 IEEE/SICE International Symposium on System Integration}, pp. 1144--1149, 2020.
\bibitem{Tsunoda_ex2}
Y. Tsunoda, L. Nghia, Y. Sueoka, and K. Osuka, ``Experimental Analysis of Shepherding-Type Robot Navigation Utilizing Sound-Obstacle-Interaction,'' {\it Journal of Robotics and Mechatronics}, Vol. 35, No. 4, pp. 957--968, 2023.
\bibitem{pierson}
A. Pierson and M. Schwager, ``Controlling Noncooperative Herds with Robotic Herders," {\it IEEE Transactions on Robotics}, Vol. 34, No. 2, pp. 517--525, 2018.
\bibitem{Boidmodel}
C. W. Reynolds: Flocks, Herds, and Schools, ``A distributed
behavioral model,'' {\it ACM SIGGRAPH Computer Graphics},
Vol. 21, No. 4, pp. 25--34, 1987.
\bibitem{TSP}
M. Bellmore and G. L. Nemhauser, ``The traveling salesman problem: A survey,''{\it Operations Research}, Vol. 16, No. 3, pp538--558, 1968.
\bibitem{WSP}
J. Bossek, K. Casel, P. Kerschke, and F. Neumann, ``The node weight dependent traveling salesperson problem: approximation algorithms and randomized search heuristics,''{\it In 2020 Genetic and Evolutionary Computation Conference}, pp. 1286--1294, 2020.
\bibitem{RLS}
F. Neumann, and W. Ingo, ``Randomized local search, evolutionary algorithms, and the minimum spanning tree problem," 
{\it Theoretical Computer Science}, Vol. 378, No. 1, pp. 32--40, 2007.
\bibitem{RLS_metamol}
S. Jens, T. Karsten, W. Ingo, ``The analysis of evolutionary algorithms on sorting and shortest paths problems,"
{\it Journal of Mathematical Modelling and Algorithms}, Vol. 3, No. 4, pp. 349--366, 2005.

\end{thebibliography}
\end{document}